\documentclass[journal]{IEEEtran}
\usepackage{cite}
\ifCLASSINFOpdf
   \usepackage[pdftex]{graphicx}
\else
\fi
\usepackage{amsmath}
\usepackage{array}
\usepackage{hyperref}

\usepackage{multirow}
\usepackage{color}
\usepackage[normalem]{ulem} % in order to use the 'delete' line

\begin{document}
\title{Fusion of Heterogeneous Earth Observation Data for the Classification of Local Climate Zones}

\author{Guichen~Zhang; 
%~\IEEEmembership{Member,~IEEE,}
Pedram~Ghamisi,~\IEEEmembership{Senior Member,~IEEE;}
Xiao Xiang~Zhu,~\IEEEmembership{Senior Member,~IEEE}
      %and~Jane~Doe,~\IEEEmembership{Life~Fellow,~IEEE}% <-this % stops a space
\thanks{This work is jointly supported by the European Research Council (ERC) under the European Union's Horizon 2020 research and innovation programme (grant agreement No. [ERC-2016-StG-714087], Acronym: \textit{So2Sat}), Helmholtz Association under the framework of the Young Investigators Group ``SiPEO'' (VH-NG-1018, www.sipeo.bgu.tum.de), and the Bavarian Academy of Sciences and Humanities in the framework of Junges Kolleg. 
(\textit{Corresponding Author: Xiao Xiang Zhu})} 
\thanks{G. C. Zhang is with the School of Remote Sensing and Information Engineering,
Wuhan University, 430079, Wuhan, Hubei, China, and also with the Remote Sensing Technology Institute (IMF), German Aerospace Center (DLR), 82234, Wessling, Germany (email: guichen.zhang@dlr.de).}% <-this % stops a space
\thanks{P.Ghamisi is with Remote Sensing Technology Institute (IMF), German Aerospace Center (DLR), 82234 Weßling, Germany; and is with Helmholtz-Zentrum Dresden-Rossendorf (HZDR), Helmholtz Institute Freiberg for Resource Technology (HIF), Exploration, D-09599 Freiberg, Germany (email: p.ghamisi@gmail.com)}
\thanks{X. Zhu is with the Remote Sensing Technology Institute (IMF), German Aerospace Center (DLR), Germany and with Signal Processing in Earth Observation (SiPEO), Technical University of Munich (TUM), Germany (e-mails: xiao.zhu@dlr.de).}
% <-this % stops a space
}


\markboth{Draft 2018}%
{Shell \MakeLowercase{\textit{et al.}}: Bare Demo of IEEEtran.cls for IEEE Journals}



\maketitle
%\textcolor{blue}{The revised parts of the paper are marked with blue color.} \\\par
\begin{abstract}
--\textit{This is a preprint. To read the final version please
visit IEEE XPlore.}

This paper proposes a novel framework for fusing multi-temporal, multispectral satellite images and OpenStreetMap (OSM) data for the classification of local climate zones (LCZs). Feature stacking is the most commonly-used method of data fusion but does not consider the heterogeneity of multimodal optical images and OSM data, which becomes its main drawback. The proposed framework processes two data sources separately and then combines them at the model level through two fusion models (the landuse fusion model and building fusion model), which aim to fuse optical images with landuse and buildings layers of OSM data, respectively. In addition, a new approach to detecting building incompleteness of OSM data is proposed. The proposed framework was trained and tested using data from the 2017 IEEE GRSS Data Fusion Contest, and further validated on one additional test set containing test samples which are manually labeled in Munich and New York.
Experimental results have indicated that compared to the feature stacking-based baseline framework the proposed framework is effective in %\sout{analyzing multi-source data fusion}
fusing optical images with OSM data for the classification of LCZs with high generalization capability on a large scale. The classification accuracy of the proposed framework outperforms the baseline framework by more than 6\% and 2\%, while testing on the test set of 2017 IEEE GRSS Data Fusion Contest and the additional test set, respectively.
%\sout{The classification accuracy of the proposed framework outperforms the winner of the 2017 IEEE GRSS Data Fusion Contest by more than 1\%.} 
In addition, the proposed framework is less sensitive to spectral diversities of optical satellite images and thus achieves more stable classification performance than state-of-the-art frameworks.
\end{abstract}


% \fi
\IEEEpeerreviewmaketitle

\section{Introduction}
\IEEEPARstart{U}{rbanization} has raised widespread concerns during the past few decades\cite{Lowry1977,Brunn2003,Lin1994}. Many urban climate models have been formed in order to study the combined effect of urban climate and climate change on urban areas and to assess the vulnerability of urban populations\cite{Bechtel2015}. It is, therefore, necessary to use a quantitative urban landscape description as the input of urban climate models\cite{Bechtel2015}. 

Most of the studies dedicated to the urban landscape description concentrated either on separating urban areas from rural areas\cite{Jin2005,Esch2013}
or generating local climates under different standards\cite{Bechtel2015}.
However, the binary schemes that separate urban areas from rural areas were not enough to characterize cities because there were many sub local climates under urban or rural categories that were nontrivial for urbanization studies\cite{Bechtel2015,Bechtel2012}. Consequently, a standardized scheme to characterize cities was lacking, making it hard to compare and combine their urbanization works on global and local scales\cite{Bechtel2015}. 

Local climate zones (LCZs) is the first classification scheme providing a generic, complete, largely comprehensive, and disjoint discretization of urban landscapes with respect to the internal physical structures of urban areas on a global scale\cite{Bechtel2015,Stewart2012}.
The LCZ scheme is based on urban functions and climate-relevant surface properties, instead of only build-ups, which are more appropriate for urban studies\cite{Stewart2012}.
Besides, it is a globally standardized and generalized scheme with inter-city comparability, and it is nonspecific to time, place, and culture\cite{Bechtel2015}.
The LCZ scheme describes urban areas in different levels of detail. This paper considers the LCZ scheme at level 0, where LCZs consist of 10 built labels and 7 landcover labels (see Fig. 1)\cite{Bechtel2015}. 
\begin{figure}[!t]
\centering
\includegraphics[width=3.5in]{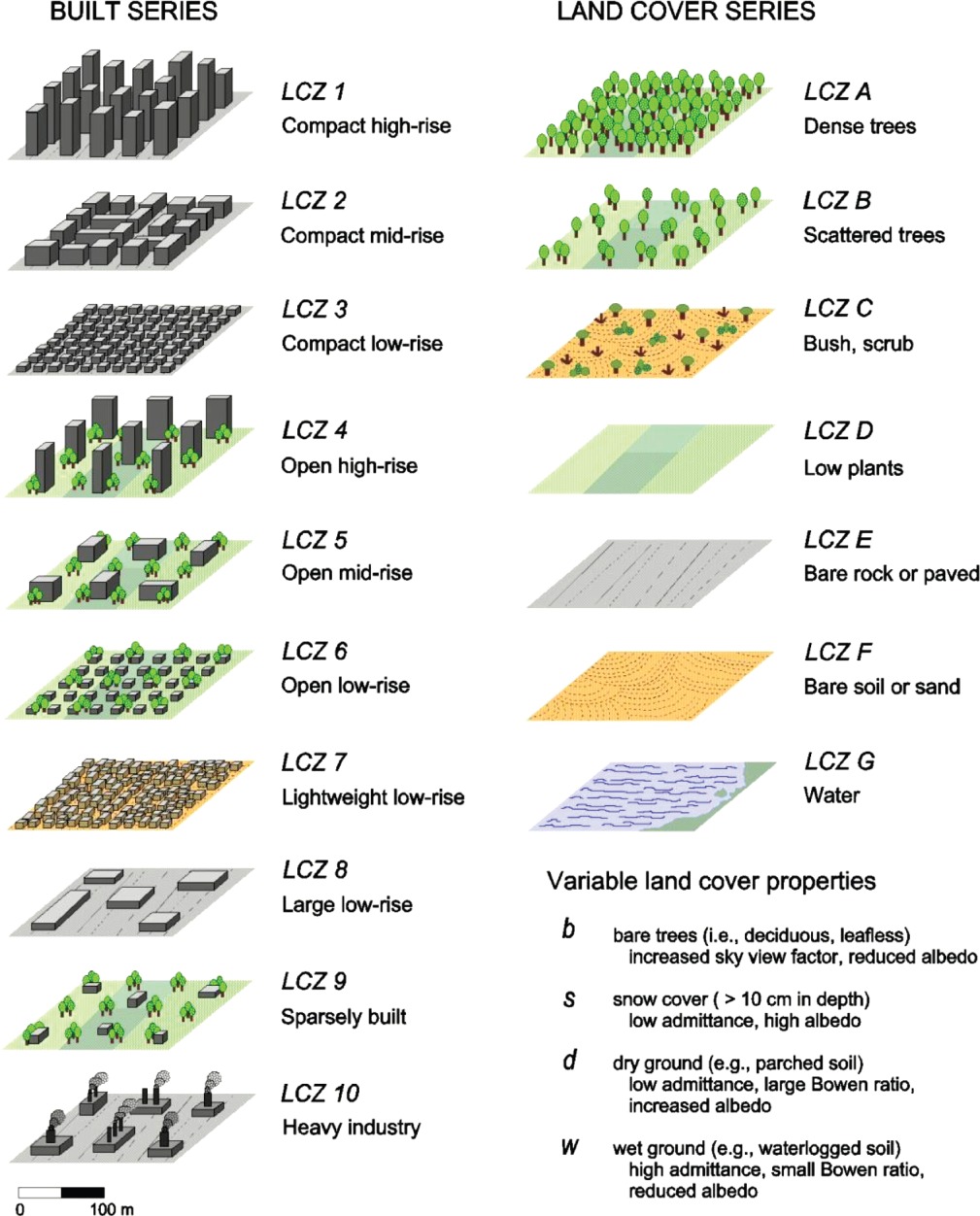}
\caption{LCZ Classification Scheme\cite{Stewart2014}
%Evaluation of the ‘local climate zone’ scheme using temperature observations and model simulations
}
\label{lczs_definition}
\end{figure}

Compared with field studies, satellites provide high spatial resolution images with continuous observations from space, offering a large potential in urban mapping. Moreover, OpenStreetMap (OSM)\cite{OpenStreetMap}
%\footnote{Map data copyrighted OpenStreetMap contributors and available from https://www.openstreetmap.org}
data have become one of the most popular free-accessible maps (https://www.openstreetmap.org), providing the effective complement of satellite images\cite{Johnson2017,Audebert2017}. % ~\parencite{johnson2017employing,audebert2017joint,schultz2017open}
Furthermore, multi-source data fusion offers much potential for urban mapping. Due to the rich characteristics of natural processes and environments, it is rare for a single acquisition method to provide a complete understanding of certain phenomenon\cite{Lahat2015,hu2018feature,qiu2018feature}.
%~\parencite{Lahat2015}.
Multi-source data fusion considers the task from various points of view and then provides opportunities to view the whole picture.
Therefore, this paper aims to fuse satellite images and OSM data for the classification of LCZs on a global scale. This involves three issues: classification, data fusion, and global mapping.

\subsection{Classification}
%, e.g. SVM, neural network, random forest,  .... on all image classification tasks. point out that in LCZs classification, we have
In the past few decades, researchers have developed many effective and efficient methods for image classification. For instance, Lu and Weng\cite{Lu2007} gave a comprehensive review and grouped classification methods in various ways, depending on supervised or unsupervised, parametric or nonparametric, hard or soft, and pixel, sub-pixel, or object based. They summarized different classification methods as the following four points\cite{Lu2007}. First, using supervised or unsupervised methods depends on whether training samples are available. Second, parametric methods assume that the data are subject to certain statistical distributions, which are often violated, especially in complex landscapes. Besides, much previous research has indicated that nonparametric classifiers may provide better classification results, compared to parametric classifiers, in complex landscapes. Third, hard classification assigns each pixel to a certain class, and soft classification gives a measure of belonging to each pixel. Furthermore, soft classification provides more information and potentially a more accurate result, especially for coarse spatial resolution data classification. Fourth, which level(s) of classification we use depends on the application. Pixel-level classification is straightforward and easier to implement, but it ignores the impact of mixed pixels. Sub-pixel level classification considers the heterogeneous information in one pixel and provides a more appropriate representation and area estimation of land covers than per-pixel approaches. Object-level classification firstly merges pixels into homogeneous areas and then classifies based on homogeneous areas.

This paper concentrates on the supervised, nonparametric, soft, and per-pixel classification method due to the following reasons.
First, training samples are available from the contest\cite{Tuia2017}. Second, previous works indicate that nonparametric and soft classification give a better classification performance, especially when classifying complicated urban scenes\cite{Lu2007}. Third, for simplicity, this paper concentrates on the pixel-level classification and ignores the heterogeneous information within one pixel. In addition, we do not use object-based classification approaches, considering the application. The difference between LCZ classification and landcover mapping\cite{Anderson1976} is that LCZ labels are defined as a certain arrangement of various objects while the labels of landcover mapping are defined as objects. The segmentation process may break the certain arrangement into several pieces in the LCZ classification, where the segmented areas may lose physical structures that are key to identifying LCZ labels. 

Some works\cite{Maxwell2018, Lu2007, Ghamisi2017, 8113122} have summarized and compared the most commonly used supervised and pixel-based classifiers, which are support vector machines (SVMs)\cite{Huang2002}, random forest (RF)\cite{Rodriguez-Galiano2012}, and neural networks (NNs)\cite{Zhu2017}.
% compare three Algs. Implementation of machine-learning classification in remote sensing: an applied review \cite{Maxwell2018}
An SVM aims to find optimal linear or non-linear boundaries in high-dimensional feature spaces with or without using kernels. It is less sensitive to smaller training sets than NNs but more sensitive to the training data quality than an RF\cite{Maxwell2018}. Additionally, its user-defined parameters are fewer than NNs\cite{Maxwell2018}. Compared with an RF, the computation burden of an SVM is larger in the presence of a large feature quantity and when using the kernel trick\cite{Maxwell2018}. 
An RF is an ensemble of many weak classifiers (decision trees). It is less sensitive to smaller training sets than NNs and less sensitive to the training data quality than SVMs\cite{Maxwell2018}. It can generate soft classification results (votes of trees), which provide more information. Furthermore, its user-defined parameters are fewer than SVMs\cite{Maxwell2018}. Recently, Rainforth and Wood\cite{Rainforth2015} proposed an improved forest method, called canonical correlation forest (CCF), which naturally embeds the correlation between input features and labels in hyperplane splits and outperforms 179 classifiers considered in a recent extensive survey paper\cite{Fernandez-Delgado2014}. Based on the studies reported in \cite{8113122}, a CCF outperforms an SVM, RF, and NNs in terms of classification accuracies for hyperspectral data. NNs are currently a popular method and aim to tune hyper-parameters of neural networks. It can achieve a quite good classification performance with well-determined conditions\cite{Zhu2017,Ghamisi2017}. An NN is usually sensitive to smaller training sets and training data quality and has severe over-fitting problems when there are not enough training samples. The classification performance highly depends on the architecture of NNs, which contains many user-defined parameters\cite{Maxwell2018,Zhu2017}. This creates a very high computation burden, especially when the network goes deeper\cite{Zhu2017}. 

This paper aims to develop an efficient and worldwide adaptive framework for the classification of LCZs. We, therefore, intend to choose one or several classifiers with less computational burden, less over-fitting, and higher transferability among different geo-locations and better robustness over noise. 
As a result, in this paper, the CCF was chosen among all types of classifiers. A more detailed description of CCF will be provided in the methodology section. 

%\textcolor{red}{why do we have the subsection "classifier" in this paper? It is a bit beside the point. We should talk about it.}
\subsection{Data Fusion}
The World Urban Database and Access Portal Tools (WUDAPT) project (http://www.wudapt.org/) was launched in 2012, with the aim of developing worldwide urban local climate mappings \cite{Mills2015}.
%~\parencite{mills2015introduction}.
It has provided a standard classification framework that generates LCZ maps by using freely available optical satellite images, such as Landsat-8, and manually selected ground truth on Google Earth.
%~\parencite{xu2017issues, WUDAPT_project}.
In addition to the use of spectral bands captured by satellite images, some frameworks have also jointly considered several data sources, such as temperature\cite{Bechtel2012,Kaloustian2017}, building height\cite{Bechtel2012,Bechtel2017}, mean amplitude of synthetic aperture radar (SAR) images\cite{Kaloustian2017}, and OSM data\cite{Lopes2017, Yokoya2017, Xu2017, Sukhanov2017, Anjos2017}.
% temperature and building height data. ~\parencite{bechtel2012classification}
% mean amplitude of the \gls{sar} image ~\parencite{kaloustian2017local} 
% 4 contest papers
Most of the above fusion studies extracted features from different data sources and then applied the feature stacking approach for data fusion as feature stacking is one of the most commonly used and fastest implemented fusion methods. Those studies assumed that classification performance improves after using more features extracted from multi-source data.
%different labels have better separability after feeding into classifiers through enlarging the feature number. 
Lopes, Fonte, See et al.\cite{Lopes2017} fused OSM data directly with LCZ maps from WUDAPT without using feature stacking.
They manually correlated the LCZ labels and OSM feature classes and then assigned the areas with typical OSM feature classes into certain LCZ labels.

Current approaches to fuse satellite images with OSM data have several limitations aroused by data heterogeneity. 
%For instance, feature stacking is not an appropriate approach to fuse satellite images with OSM data because these two data sources are heterogeneous.
%First, Satellite images and OSM data have totally different kinds of acquisition techniques. Satellite images records ground information from space observations whereas OSM data record features of geometric units (e.g. point, line, and polygon) from local experts.
First, satellite images and OSM data have different kinds of acquisition techniques. Satellite images are recorded by space observations whereas OSM data are recorded by local experts.
Second, satellite images and OSM data have different data forms and spatial resolutions. Satellite images are raster data with limited resolutions (10 meters to 100 meters, in this paper) whereas OSM data are in vector format, which can be rasterized into any resolutions. The differences of data form bring many difficulties in data fusion. On one side, downsampling the OSM data into the resolution of satellite images results in the lose of much valuable information on OSM; alternatively, upsampling all satellite images will significantly increase the computational burden without adding any useful information. 
Third, satellite images and OSM data have different noise sources. The noise sources of satellite images come from the imaging chain (e.g., satellite platform vibration, atmosphere, etc.) whereas the noise sources of OSM data are created by the individuals recording the data, causing OSM data to contain errors or incomplete recordings.
%~\parencite{lahat2015multimodalfusionreview}
The approach to manually correlating OSM and LCZ maps\cite{Lopes2017} could somehow resolve the data heterogeneity problem, but it needs human labor to consider the correlation, which costs much time and money. The other drawback is that feature classes of OSM and LCZ labels follow different classification schemes. There is the minor possibility that some areas with certain OSM features could be directly assigned to certain LCZ labels. This demonstrates the necessity of forming novel approaches to resolve issues with heterogeneous data fusion; this is a topic deeply investigated in this paper.
%Thus, it is highly needed to derive novel approaches to deal with heterogeneous data fusion which is deeply investigated in this paper.
%parencite{lopes2017using}
%Moreover, anomaly detection methods...

\subsection{Global Mapping}
%~\parencite{kaloustian2017local}
Global LCZ mapping assists greatly in studying and comparing local climates on regional and worldwide scales. 
Satellite images are influenced by diverse spectral information due to complicated physical procedures in the imaging chain. This spectral diversity could decrease the classification performance, especially when analyzing multi-temporal, multispectral, and multi-location classification\cite{Tuia2016}.
Many studies have successfully generated LCZ maps of one city by labeling samples of that city, and they have achieved satisfactory classification performance (e.g., overall accuracies (OAs) were beyond 80\%). Meanwhile, it is of great interest to train models from the samples of some cities and apply the models to other cities since it costs much time and human labor to label all cities worldwide.
One study \cite{Kaloustian2017} tried to select training samples from one city for the classification of another city by using an RF. The classification accuracies dropped to 18.2\%, which indicates that the knowledge transferability between different cities should be carefully considered. 

Thanks to the 2017 IEEE GRSS Data Fusion Contest\cite{Tuia2017} organized by IEEE Geoscience and Remote Sensing Society, some promising works have been accomplished in multi-model remote sensing data fusion and their transferability studies in the application of LCZ classification. %parencite{tuia20172017}
% introduce four works on how they deal with inter-cities adaptive problem 
The contest provided training samples from five cities and test samples from four other cities. Four novel frameworks with achieved top classification accuracies were selected; their works are quite promising and intriguing. 
%{igrss_lcz_multimultimulti}
Yokoya et al. \cite{Yokoya2017} introduced CCFs\cite{Rainforth2015} in analyzing knowledge transferability between cities and received the best result (OA was 74.94 \%) in the contest. A CCF is an advanced forest classifier that naturally incorporates both the labels and the correlation between the input features in the choice of projection for computing decision boundaries in the projected feature space. Results demonstrate that the CCF has much better performance than other forest classifiers, such as the RF\cite{Rodriguez-Galiano2012} and rotation forest (RoF)\cite{Rodriguez2006}, when the training and test samples are not from the same domain\cite{Yokoya2018}. Besides CCFs, three more works have managed to approach the intercity transferability problem by 
%Instead of hoping the extracted features of training samples are generalized enough to represent the extracted features of testing samples, CCF computes hyperplane splitting in a projected space through canonical correlation analysis\cite{Hotelling1936}. It embeds the correlation between features and labels, thus boosting the classification performance\cite{Rainforth2015}. Besides CCF, three more works managed to deal with inter-cities transferability problem by projectthefeaturesinto 
%igrss_lcz_cotraining, igrss_lcz_multilevelensemble,
developing a co-training process\cite{Xu2017}, 
%igrss_lcz_cotraining
ensembling various classifiers\cite{Sukhanov2017}, 
% igrss_lcz_multilevelensemble
and conducting object-based classification\cite{Anjos2017}
% igrss_lcz_urbanenvironments
approaches. The best OAs that they have achieved are 73.2\%,
%igrss_lcz_cotraining
72.63\%,
%igrss_lcz_multilevelensemble
and 72.38\%, 
%igrss_lcz_urbanenvironments
respectively. 
However, the spectral diversity between training and test samples still plays a large impact on those frameworks. Therefore, further studies are needed on creating a generalized LCZ classification framework with more stable behavior. 

Accordingly, this paper proposes a novel framework of fusing satellite images and OSM data for the classification of LCZs on a global scale. First, we extract features from satellite images and OSM data, respectively, and analyze these two features separately. Second, we apply different models to the extracted features from the satellite images and OSM data instead of applying a simple stacking of those two features. In addition, we propose a simple yet effective approach to detect the areas with incomplete recordings in the OSM data. Finally, we fuse the results from different models by conducting a weighting process. The main contributions of this paper are thus as follows:
\begin{enumerate}
\item The proposed framework analyzes the heterogeneity between satellite images and OSM data, and we conclude that current frameworks based on feature stacking have many limitations.
\item This paper proposes a novel idea of fusing satellite images and OSM data by taking the data heterogeneity into account. In this context, instead of simply stacking the heterogeneous features, we apply different models to various data modality\cite{Gomez2015} in a separate manner and then conduct a novel fusion approach.
\item The proposed fusion approaches achieve a robust classification performance on a global scale by carefully fusing OSM with satellite images.
\item We propose a novel approach to detect building data incompleteness by considering the correlation between buildings and the landuse layers of OSM.
\end{enumerate}

The remaining portion of this paper is organized as follows. In Section II, we introduce the dataset, study regions, and data preprocessing. In Section III, we introduce the proposed framework. In Section IV, we define the baseline framework and compare its classification performance with the proposed framework. In addition, we also present the feature importance rankings and the effectiveness of the approach of detecting incomplete building recordings. Finally, we conclude our work and give future directions in Section V.

\section{Dataset}
\subsection{Data Fusion Contest Dataset}
%This paper uses the dataset identified as "grss\_dfc\_2017"\cite{grss_dfc_2017}, made freely available by the 2017 IEEE GRSS Data Fusion Contest\cite{Tuia2017}. A detailed data description can be found in \cite{dataformatinfo_contest}.
The dataset of the data fusion contest (DFC)\footnote{The data fusion contest (DFC) dataset refers to 2017 IEEE GRSS Data Fusion Contest\cite{Tuia2017} unless otherwise noted.} identified as "grss\_dfc\_2017"\cite{grss_dfc_2017} was made freely available by the 2017 IEEE GRSS Data Fusion Contest\cite{Tuia2017}. A detailed data description can be found in \cite{dataformatinfo_contest}. 
The data consist of training samples selected from five cities (Berlin, Hong Kong, Paris, Rome, and Sao Paulo) and test samples selected from four cities (Amsterdam, Chicago, Madrid, and Xi'an) (see Fig. 2). In each city, the data contain multi-temporal Landsat-8 images, single-temporal Sentinel-2 images, and OSM data. Satellite images are of 1C-level and have 100 meter spatial resolution. OSM data include buildings, landuse, water, and natural layers, which are available in both raster and vector data formats. The raster form of OSM data is of 5 meter spatial resolution. A buildings layer is a binary layer delineating building areas. A landuse layer separates an area into different landuse classes\cite{ramm2014}.\footnote{Landuse layers may contain the following classes: forest, park, residential, industrial, farm, cemetery, allotments, meadow, commercial, nature\_reserve, recreation\_ground, retail, military, quarry, orchard, vineyard, scrub, grass, heath, and national\_park.} OSM data also include road layers, which are only available in the vector data form. 
Moreover, the distribution of training and test sizes are provided in Table I.

In this paper, we redownloaded Landsat-8 images from the U.S. Geological Survey (https://earthexplorer.usgs.gov/) and Sentinel-2 images from the Sentinel Data Hub (https://scihub.copernicus.eu/) to acquire the original spatial resolution images. Then, atmospheric corrections were conducted and cloud masks were generated. We only kept those redownloaded images with exactly the same geo-location and time acquisition as grss\_dfc\_2017\cite{grss_dfc_2017}. In this work, buildings and landuse layers from OSM data were only used because it was discovered that the water and natural layers were not available for all cities. In addition, road layers were not considered in this work. 

Data preprocessing was conducted on Landsat-8 images, Sentinel-2 images, the buildings layers of OSM, and the landuse layers of OSM. 
For Landsat-8 images, atmospheric corrections using ATCOR-2/3 version 9.0.0 with the haze removal option were conducted, and then the data were upsampled into 10 meter spatial resolution using bicubic interpolation. 
For Sentinel-2 images, atmospheric corrections were conducted using Sen2Cor version 2.3.1, and then the data were upsampled  into 10 meter spatial resolution using bicubic interpolation. After preprocessing, Landsat-8 images contained the Bands 1, 2, 3, 4, 5, 6, 7, 8, 10, and 11, and Sentinel-2 images contained the Bands 2, 3, 4, 5, 6, 7, 8, 8A, 11, and 12. We did not consider cirrus and water vapor bands because they are mainly dedicated to cirrus detections and water vapor corrections and are not usually used in urban mapping\cite{wang2016}. We also did not consider coastal/aerosol bands of Sentinel-2 because they are dedicated to aerosol retrieval and cloud detection\cite{wang2016}. Besides, a cloud mask was generated from satellite images at each acquisition time. The areas that contained high cloud probability were removed in the following process.
% ~\parencite{wang2016fusion}
For buildings and landuse layers of OSM data in raster form, each layer was normalized between 0 and 1. For the buildings layers of OSM data in vector form, the layers of building central points were extracted using ArcMap version 10.5.1, and then the layers of building central points were rasterized into 5 meter spatial resolution. 

\subsection{Additional Test Dataset}
In addition to the four test cities available in grss\_dfc\_2017\cite{grss_dfc_2017}, we have used an additional test (AT) dataset by selecting two extra test cities (Munich and New York, see Fig. 2) to validate our proposed framework.
Ground truth data were labeled according to the LCZ classification scheme\cite{Stewart2014}. Table I shows the distribution of test sizes.
In each city, we have downloaded multi-temporal Sentinel-2 images from the Sentinel Data Hub (https://scihub.copernicus.eu/) and OSM data from Geofabrik (https://www.geofabrik.de/). After that, Sentinel-2 images and OSM data were processed according to the strategy in II.A.

\begin{table}[!t]
\newcommand{\tabincell}[2]{\begin{tabular}{@{}#1@{}}#2\end{tabular}}
%% increase table row spacing, adjust to taste
\renewcommand{\arraystretch}{1.3}
% if using array.sty, it might be a good idea to tweak the value of
% \extrarowheight as needed to properly center the text within the cells
\caption{Training and Test Samples}
\label{extracted_features}
\centering
\begin{tabular}{cccc} % |c||c|
\hline
Lable No. & Training Size & Test Size (DFC) & Test Size (AT)\\
\hline
1 & 1642 & 242 & 526\\
2 & 6103 & 4892 & 1414\\
3 & 5738 & 1535 & 2825\\
4 & 2098 & 2270 & 172\\
5 & 4759 & 2255 & 1131\\
6 & 8891 & 8265 & 4271\\
7 & 0 & 0 & 0\\
8 & 4889 & 11230 & 2304\\
9 & 1156 & 1072 & 3763\\
10 & 449 & 920 & 887\\
11 & 17716 & 3170 & 2175\\
12 & 2819 & 4528 & 362\\
13 & 1741 & 1284 & 41\\
14 & 14457 & 12994 & 2949\\
15 & 323 & 1104 & 253\\
16 & 503 & 391 & 0\\
17 & 8561 & 4454 & 12626\\
\hline
\end{tabular}
\end{table}

\begin{figure}[!t]
\centering
\includegraphics[width=3.5in]{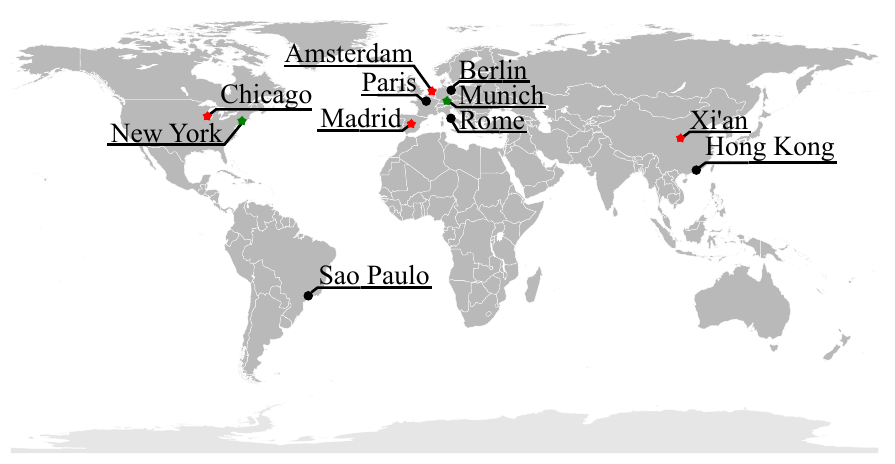}
% where an .eps filename suffix will be assumed under latex, 
% and a .pdf suffix will be assumed for pdflatex; or what has been declared
% via \DeclareGraphicsExtensions.
\caption{Training cities (Berlin, Hong Kong, Paris, Rome, and Sao Paulo) marked with black dots from grss\_dfc\_2017\cite{grss_dfc_2017}, test cities (Amsterdam, Chicago, Madrid, and Xi'an) marked with red dots from grss\_dfc\_2017\cite{grss_dfc_2017}, and additional test cities (Munich and New York) marked with green dots from additional test dataset.}
\label{dataset_map}
\end{figure}

\section{Methodology}
In this section, we propose a novel framework to fuse satellite images with OSM data (see Fig. 3). First, spectral, spatial, textural, and map features were extracted from satellite images and OSM data.
%satellite features are extracted, including: building density features and landuse features from satellite images, building central points layers and landuse layers, respectively.
Second, three different models were applied to these three kinds of extracted features. Specifically, CCFs\cite{Rainforth2015} were applied to the satellite features, and then a landuse fusion model and building fusion model were derived to fuse landuse features and building density features with satellite features. In addition, a novel approach was also proposed to mask out incomplete building areas. Finally, postprocessing and decision fusion were conducted. 

\begin{figure*}[!t]
\centering
\includegraphics[width=7in]{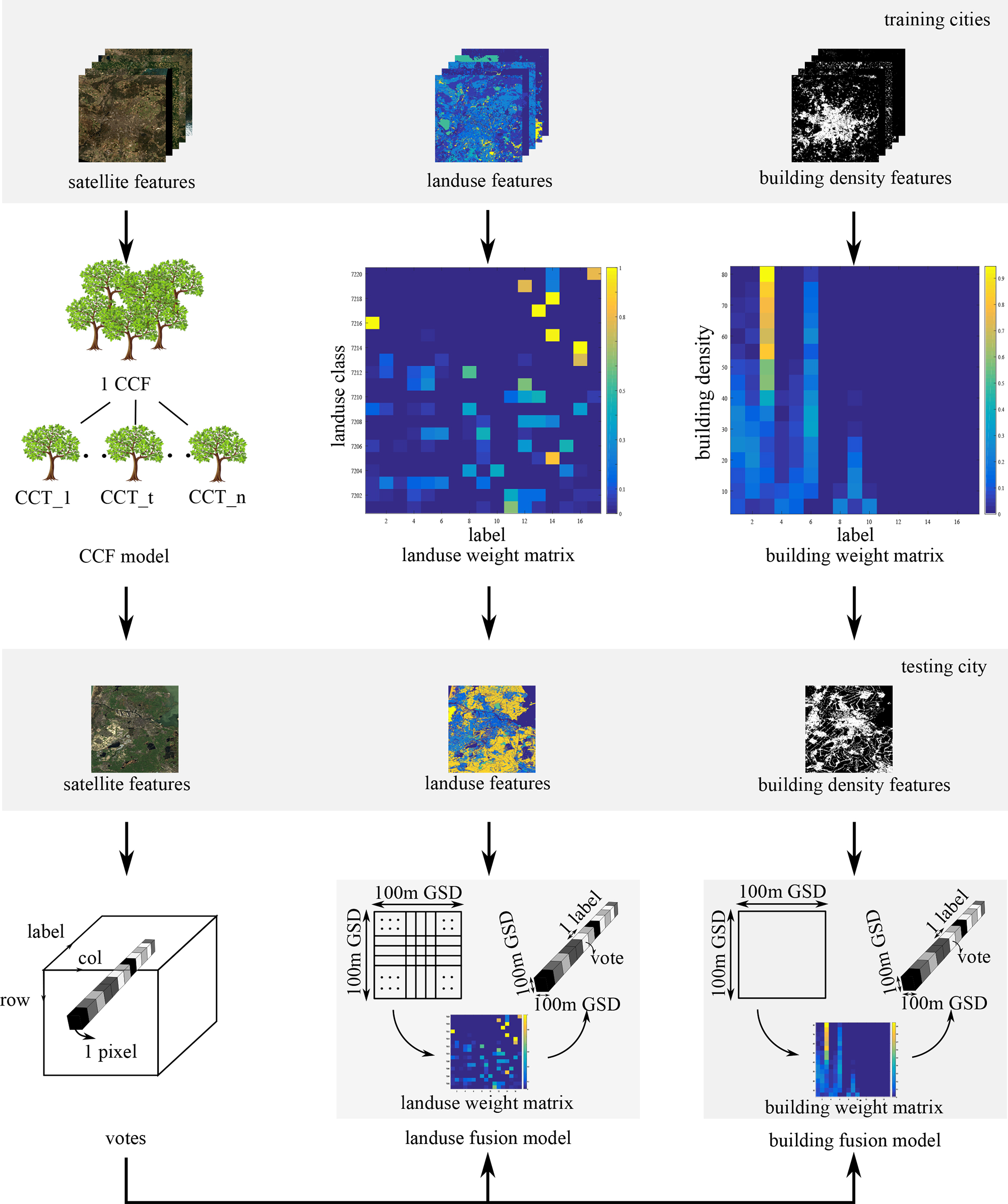}%
\caption{Proposed framework.}
\label{propose_framework}
\end{figure*}

\subsection{Feature Extraction}
Spectral, spatial, and texture features were extracted from the preprocessed Landsat-8 and Sentinel-2 images through the same computation process at each acquisition time. First, mean values and standard deviations of all bands of images in each patch of 100 $\text{m}$ ground sample distance (GSD)
%$\text{m}^{2}$
were computed. Second, three spectral indexes (normalized difference vegetation index (NDVI), normalized difference water index (NDWI), and bare soil index (BSI)) were derived, and then their mean values and standard deviations were computed in each patch of 100 $\text{m}$ GSD. Third, the mean values of morphological profiles (MPs) of NDVI were computed in each patch of 100 $\text{m}$ GSD. Fourth, a weighted gray-level co-occurrence matrix (GLCM) algorithm\cite{zhang2017} was used to produce contrast, correlation, energy, and homogeneity texture features, and then their mean values were computed in each patch of 100 $\text{m}$ GSD.
Besides, building density features were extracted from the layers of building central points by counting the building number in each patch of 100 $\text{m}$ GSD. In addition, the preprocessed landuse and buildings layers were also used as the landuse and building features.  

Table II lists the extracted features' names, quantities, and spatial resolutions. Since the feature extraction from Landsat-8 or Sentinel-2 images shares the same computation process and those extracted features from the two satellites have the same feature names, the spectral, spatial, and textural features in Table II represent the extracted features from either Landsat-8 or Sentinel-2 images. 
Moreover, the spectral, spatial, and textural features from either Landsat-8 or Sentinel-2 images are named as satellite features. 
%Moreover, we name the extracted features from Landsat-8 images as the Landsat-8 features, and we name the extracted features from Sentinel-2 images as the Sentinel-2 features. For simplicity, we use image features to represent either Landsat-8 features or Sentinel-2 features. 
%Moreover, we group the extracted features from either Landsat-8 or Sentinel-2 and name them as 'satellite features', because they have the same fusion process with OSM data. The fusion of Landsat-8 and Sentinel-2 data only happens at the decision level. 

\begin{table}[!t]
\newcommand{\tabincell}[2]{\begin{tabular}{@{}#1@{}}#2\end{tabular}}
%% increase table row spacing, adjust to taste
\renewcommand{\arraystretch}{1.3}
% if using array.sty, it might be a good idea to tweak the value of
% \extrarowheight as needed to properly center the text within the cells
\caption{Extracted Features}
\label{extracted_features}
\centering
%% Some packages, such as MDW tools, offer better commands for making tables
%% than the plain LaTeX2e tabular which is used here.
\begin{tabular}{ccc} % |c||c|
\hline
Feature Name & Quantity & Spatial Resolution\\
\hline
Mean values of bands & 10 & 100\\
\tabincell{c}{Mean values of\\NDVI, NDWI and BSI\\} & 3 & 100\\
Std values of bands & 10 & 100\\
\tabincell{c}{Std values of\\NDVI, NDWI and BSI\\} & 3 & 100\\
\tabincell{c}{Contrast, Correlation,\\Energy and Homogeneity} & 4 & 100\\
MPs of NDVI & 6 & 100\\
Building density & 1 & 100\\
Building & 1 & 5\\
Landuse & 1 & 5\\
All & 39 \\
\hline
\end{tabular}
\end{table}

\subsection{Canonical Correlation Forest (CCF)}
CCF\cite{Rainforth2015} is an ensemble model based on oblique decision trees. Compared with an RF, which computes the hyperplane splits in the coordinate system of input features, a CCF naturally constructs a projected feature space by considering the correlation between input features and their corresponding labels\cite{Rainforth2015}. It is more robust to the rotation, translation, and global scaling of the input features\cite{Rainforth2015}. One CCF model is composed of many sub-models named canonical correlation trees (CCTs). One CCT is a binary decision tree with many sequential divisions and is regarded as the smallest predictive unit in this paper. Each CCT is trained independently, and the ensemble of CCTs can simultaneously improve predictive performance and provide regularization against over-fitting\cite{Rainforth2015}.
%~\parencite{rainforth2015canonical}
This section applies a CCF to the stacked satellite features (see Fig. 4) and then generates an initial classification result of a test city. The classification result from one CCF could be integrated into a $votes$ cube. The first two dimensions are the spatial dimensions (row and column directions), which have the corresponding pixel coordinates for satellite features of the test city. The third dimension has the same length with LCZ labels, which is 17. $votes(i,j,l)$ record the votes number of the $l$-the label in the pixel coordinate $(i,j)$ of a test city. The larger the votes number is, the more convincing that the pixel belongs to that label. 

\begin{figure}[!t]
\centering
\includegraphics[width=3.5in]{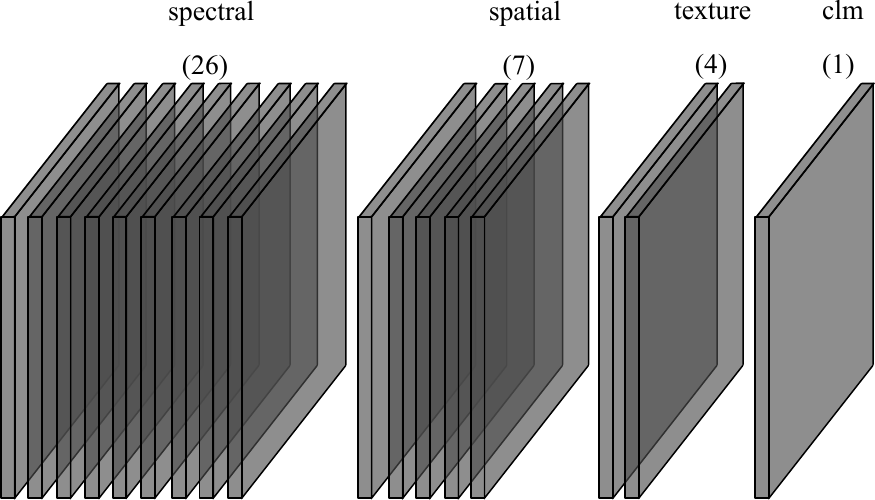}
\caption{Feature stacking of satellite features.}
\label{fea_stack}
\end{figure}

\subsection{Landuse Fusion Model}
%\IEEEpubidadjcol
A landuse fusion model contains two parts. In the first part, the relation between landuse classes and ground truth data is trained. In the second part, this relation is used as an active aid to fuse landuse features with satellite features. 

A landuse feature is denoted as $landuse(i,j)=lu$. $(i,j)$, which represents the pixel coordinates. $lu$ records the value of the pixel $(i,j)$. Those pixels where $landuse(i,j)=0$ were removed in advance.
First, landuse features and ground truth data from the training set were used to compute the prior knowledge, which was a 2D probability distribution matrix named the landuse weight matrix. The row direction of the matrix identifies different landuse classes, and the column direction identifies LCZ labels. Each element in the matrix records how large the probability is that each landuse class belongs to each LCZ label. The sum of each row is equal to 1. This matrix is denoted as $lu\_wn(lu,label)$, where $lu$ represents different landuse classes and $label$ represents different LCZ labels. 
$lu\_wn(lu, label)$ is then fused with the $votes$ cube generated from the CCF by using a weighting procedure. 
For the landuse feature $landuse$ of a test city and its corresponding $votes$ cube, the weighting procedure was conducted in each patch of 100 $\text{m}$ GSD. Since the spatial resolution of $landuse$ is 5 meters and the spatial resolution of $votes$ is 100 meters, one pixel in $votes$ corresponds to $20\times 20$ pixels in $landuse$. For each 100 $\text{m}$ GSD, the weighting procedure was conducted using the formula (1): 

\begin{equation}\label{landuse_accumu_weight1}
\begin{split}
% split 环境用\和& 来分行和设置对齐位置
luwn\_votes(i,j,:)=\sum_{d=1}^{400} lu\_wn(lu(d),:) \cdot votes(i,j,:),
\end{split}
\end{equation}
where $lu(d) \neq 0$, $(i,j)$ are the pixel coordinates in the spatial domain of $votes$, and $luwn\_votes(i,j,:)$ records the weighted $votes$ vector of pixel $(i,j)$.

\subsection{Building Fusion Model}
The building fusion model also contains two parts. In the first part, the relation between building density values and ground truth data is trained. In the second part, this relation is used as an active aid to fuse building density features with satellite features. 

A buildings layer is denoted as $build(i,j)=b$. $(i,j)$ represent the pixel coordinates. $b$ records the value of the pixel $(i,j)$ and is either $0$ or $1$.
First, building density features and ground truth data from the training set were used to compute the prior knowledge, which was a 2D probability distribution matrix named the building weight matrix. The row direction of the matrix identifies different building density ranges, and the column direction identifies LCZ labels. Each element in the matrix records how large the probability is that each building density range belongs to each LCZ label. The sum of each row is also equal to 1. 
Assuming the largest building density value from training samples was $bn\_max$, the building density ranges $bu$ were defined according to formula (2): 

\begin{equation}\label{range}
[0, gap], [gap+1,2gap],\cdot \cdot \cdot ,[bn\_max+1,bn\_max+gap],
\end{equation}
where $gap$ is an empirical value equal to 5 in this paper.

We use the building density ranges, instead of each building density value, because the building density values from training samples cannot cover all values from 0 to $bn\_max$ due to the limited number of training samples.
%with the step of 1 

The building weight matrix is denoted as $bu\_wn(bu, label)$, where $bu$ represents different building density ranges and $label$ represents different LCZ labels. $bu\_wn(bu, label)$ is then fused with the $votes$ cube generated from the CCF by using a weighting procedure.
For the building density feature $build$ of a test city and its corresponding $votes$ cube, the weighting procedure was conducted on each patch of 100 $\text{m}$ GSD. Since $build$ and $votes$ have the same spatial resolution (i.e., 100 meters), the weighting procedure could be directly computed for each 100 $\text{m}$ GSD according to formula (3): 
\begin{equation}\label{build_weight1}
\begin{split}
% split 环境用\和& 来分行和设置对齐位置
buwn\_votes(i,j,:) = bu\_wn(bu,:) \cdot votes(i,j,:)
\end{split}
\end{equation}
where $(i,j)$ are the pixel coordinates in the spatial domain of $votes$, and $buwn\_votes(i,j,:)$ records the weighted $votes$ vector of pixel $(i,j)$. 

One remaining problem still existed before using the building fusion model. Since the building density value is computed by counting the number of buildings in a local area, the building density value could be completely wrong if the buildings features have incomplete data recordings in that area. Therefore, it is necessary to generate building confidence masks in order to automatically mask out those areas with incomplete recordings of building data. 

\subsection{Building Confidence Mask Generation}
This paper proposes a novel approach to compute building confidence in a local area by jointly considering landuse and building features under 5 meter spatial resolution without using satellite images (see Fig. 5). 
The idea is that at least one building pixel should be near the local area of the pixel, which indicates a high building probability. This approach consists of four steps that will be discussed in the following subsections.

\begin{figure}[!t]
\centering
\includegraphics[width=3.5in]{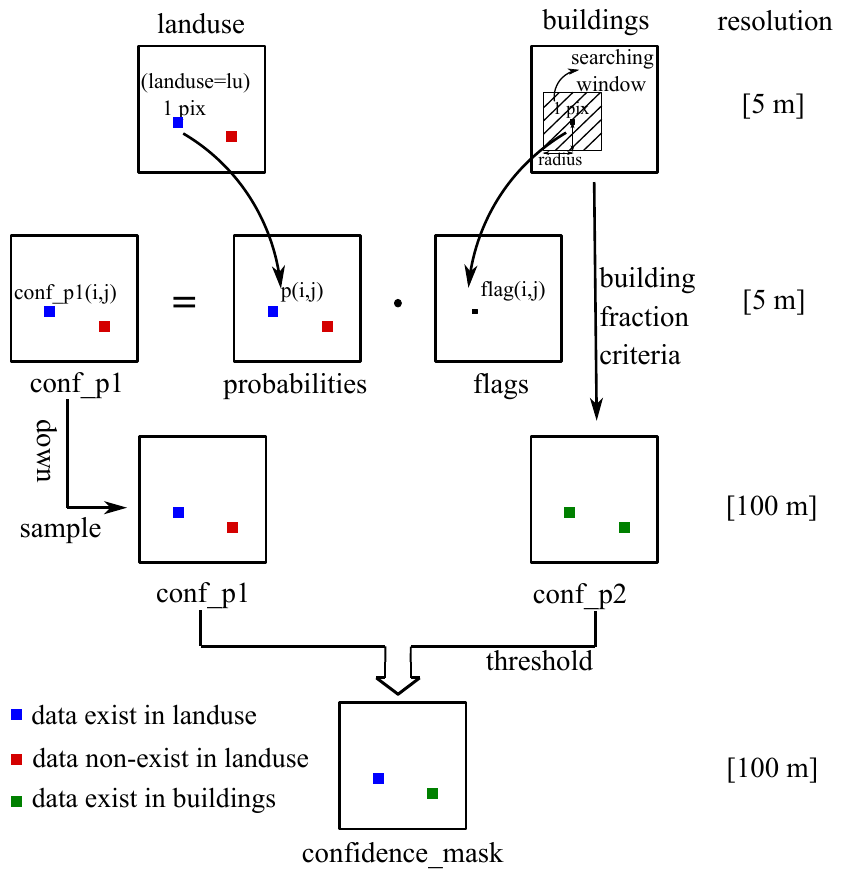}
\caption{Building confidence mask generation.}
\label{build_conf_gen}
\end{figure}

\subsubsection{Building Probability Generation}
Landuse and building features from the OSM data contain correlated knowledge. This step generated a 2D probability distribution matrix (see Fig. 6) recording the relation between landuse and building features by using both training and test samples. Zero values in landuse features were removed in advance. The row direction of the matrix identifies two different building classes: building and non-building. The column direction identifies different landuse classes. Each element in the matrix records how large the probability is that each landuse class belongs to each building class. The second row of this matrix is denoted as $p(build=1|landuse=lu)$, which is used in this approach. 
Fig. 6 presents some hints about this relation. For example, the residential class in the landuse layers has a quite high probability of being a building pixel in the buildings layers whereas the forest class in the landuse layers has a quite low probability of being a building pixel in the buildings layers.

\begin{figure}[!t]
\centering
\includegraphics[width=3.5in]{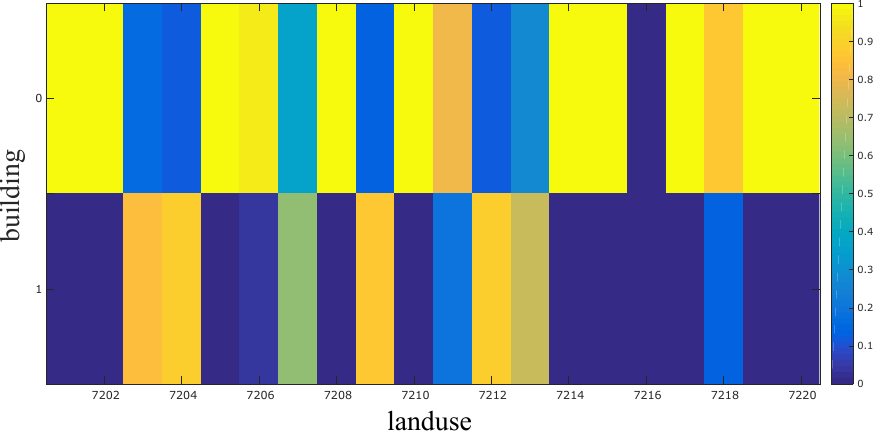}
% where an .eps filename suffix will be assumed under latex, 
% and a .pdf suffix will be assumed for pdflatex; or what has been declared
% via \DeclareGraphicsExtensions.
\caption{Building-landuse probability distribution.}
\label{build_landuse_prob}
\end{figure}

\subsubsection{Local Searching}
After generating the relation matrix, local searching was then conducted in the buildings feature of each city. The size of the searching area was given empirically after considering building intervals and is equal to 5 pixels in this paper.
For each pixel in a landuse feature, we searched building pixels around that pixel in the corresponding building feature. We used a binary value $flag$ to record the searching result. The building confidence $conf\_p1$ was computed according to: 

\begin{equation}\label{build_lu_phase1}
conf\_p1=p(build=1|landuse=lu)\cdot flag,
\end{equation}
where
\begin{equation}\label{build_lu_phase1_flag}
  flag=\left\{
   \begin{aligned}
   1&,\qquad searching \quad succeeded\\
   -1&,\qquad searching \quad failed.\\
   \end{aligned}   
  \right.
\end{equation}
 
\subsubsection{Detection Complement}
The previous step is not applicable in those pixels where the values in landuse layers are zero. Therefore, this step, which is independent of the previous step, aims to provide supplementary information when the previous step cannot be applied.
Thanks to the quantitative standard from the LCZ definition\cite{Stewart2012}, each label of LCZs defines a range of building surface fractions (see Table III). 
%\textcolor{blue}{10 \% ?? 50 \%, proves that this parameter is not sensitive! }

\begin{table}[!t]
%% increase table row spacing, adjust to taste
\renewcommand{\arraystretch}{1.3}
% if using array.sty, it might be a good idea to tweak the value of
% \extrarowheight as needed to properly center the text within the cells
\caption{Building Surface Fraction of LCZs\cite{Stewart2012}}
\label{build_surface_frac}
\centering
%% Some packages, such as MDW tools, offer better commands for making tables
%% than the plain LaTeX2e tabular which is used here.
\begin{tabular}{cc} % |c||c|
\hline
Label & Building Surface Fraction (\%)\\
\hline
Compact high-rise & 40-60\\
Compact mid-rise & 40-70\\
Compact low-rise & 40-70\\
Open high-rise & 20-40\\
Open mid-rise & 20-40\\
Open low-rise & 20-40\\
Lightweight low-rise & 60-90\\
Large low-rise & 30-50\\
Sparsely built & 10-20\\
Heavy industry & 20-30\\
Dense trees & 0-10\\
Scattered trees & 0-10\\
Bush, scrub & 0-10\\
Low plants & 0-10\\
Bare rock or paved & 0-10\\
Bare soil or sand & 0-10\\
Water & 0-10\\
\hline
\end{tabular}
\end{table}

The building surface fraction was computed in each patch of 100 $\text{m}$ GSD from building features. Then the empirical value of 10\% was used to compute the building confidence value $conf\_p2$. $conf\_p2 = 1$ if the building surface fraction was higher than 10\%; otherwise, $conf\_p2(i,j)=0$. 10\% was used as the threshold for two reasons. First, 10 \% is the lowest boundary of building surface fractions among all built types, so that sparsely built can be kept. Second, this parameter was not sensitive to the final result, which will be explained in detail in the experimental part.\\\par

\subsubsection{Combination}
The local searching result $conf\_p1$ is combined with the building fraction result $conf\_p2$ according to formula (6):
\begin{equation}\label{build_detect_combine}
  conf\_comb(i,j)=\left\{
   \begin{aligned}
   conf\_p2(i,j)&,\qquad landuse(i,j)=0 &\\
   & \ \qquad conf\_p2(i,j) = 1 &\\  
   conf\_p1(i,j)&,\qquad otherwise&\\
   \end{aligned}
  \right.
\end{equation}

Afterwards, we used an empirical value (e.g., 0.8) to threshold $conf\_comb$ into a binary layer, where $conf\_comb=1$ identified building confident areas and $conf\_comb=0$ identified building unconfident areas. 

% or


\subsection{Postprocessing and Decision Fusion}
The above process could generate one weighted $votes$ cube at each satellite acquisition time for each satellite in each test city. Therefore, if one has $T$ acquisition times for Landsat-8 images and one acquisition time for Sentinel-2 images of one test city, then $T+1$ weighted $votes$ cubes can be obtained.
Next, $T+1$ classification maps could be computed by selecting the label with the largest votes number per pixel. Then, the median filter was applied with the size of [3,3] to those classification maps. Finally, decision fusion was conducted among the $T+1$ classification maps through majority voting. 

\section{Experiment}
In this section, we firstly introduce a baseline framework and compute the contributions of the extracted features. Then, the classification performance of the proposed framework and the baseline framework is compared according to the classification accuracies and framework transferability. Afterwards, another experiment demonstrates the effectiveness of the proposed approach in generating building confidence masks.

% In addition, we also explain that the detection result is not sensitive to the threshold of building surface fraction.
The following parameters were set empirically in the experiments:
\begin{enumerate}
\item MPs: A disk-shaped structuring element was used whose sizes were 1,4,7, and 10.
\item GLCM: The number of gray levels was 32. Directions were 0$^{\circ}$, 45$^{\circ}$, 90$^{\circ}$, and 135$^{\circ}$. The offset that defined the distance of the spatial adjacency was 1 pixel. 
\item CCF: The number of CCTs was 20 to follow the literature in\cite{Yokoya2017}.
%\textcolor{red}{[You can cite our paper and say "to follow the literature". Otherwise, the reviewers think that you came up with this number in a trial and error way].
\item Building fusion model: The interval of building densities in the building weight matrix was 5. The radius of the local building searching was 25 meters GSD. The threshold of the building surface fraction was 10\%. The threshold of generating building confidence masks was 0.8.
\end{enumerate}

\subsection{Baseline Framework and Feature Importance}
The baseline framework was used to evaluate the performance of the proposed classification framework. The baseline framework directly stacks the features in Table II and feeds those features into the CCF. The building and landuse features with 5 meter spatial resolution were firstly down-sampled into 100 meter spatial resolution using the "nearest neighbor" before feature stacking. Two groups were considered where OSM data were stacked with Landsat-8 and Sentinel-2 images. Therefore, the baseline and the proposed frameworks are comparable, and the difference in their classification performance was aroused by the feature fusion models. 
In addition, feature importance was computed by using training samples through a 5-fold cross-validation to conclude which features contributed more to the classification performance. Fig. 7 illustrates the contributions of the extracted features from Landsat-8 and OSM data. Fig. 8 demonstrates the contributions of the extracted features from Sentinel-2 and OSM data. 

These two feature importance maps share many similar characteristics. First, NDVI and its MPs, which contain the information of vegetation abundance and its spatial information, obviously have higher importance than other features. The quantitative properties of LCZs could somehow explain why NDVI and its MPs are quite important here. Table IV illustrates the pervious and impervious surface fractions of LCZ labels, indicating that vegetation abundance is related to built types. For example, the compact high-rise, compact mid-rise, and compact low-rise have different ranges of pervious and impervious surface fractions in spite of the fact that they all belong to the compact-built. Therefore, although vegetation abundance could not make the built types completely separable, it still played an important role in separating different built types.

\begin{figure}[!t]
\centering
\includegraphics[width=3.5in]{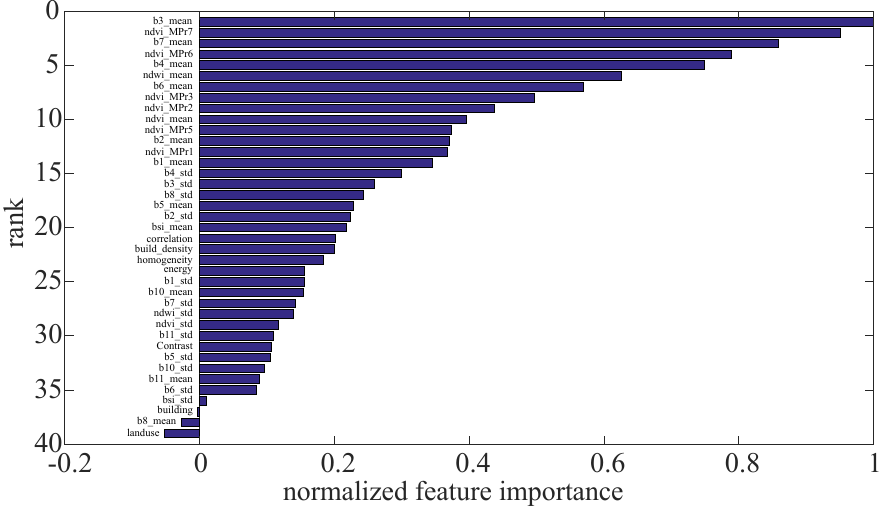}
\caption{Feature importance of the Landsat-8 group.}
\label{fea_importance_landsat}
\end{figure}

\begin{figure}[!t]
\centering
\includegraphics[width=3.5in]{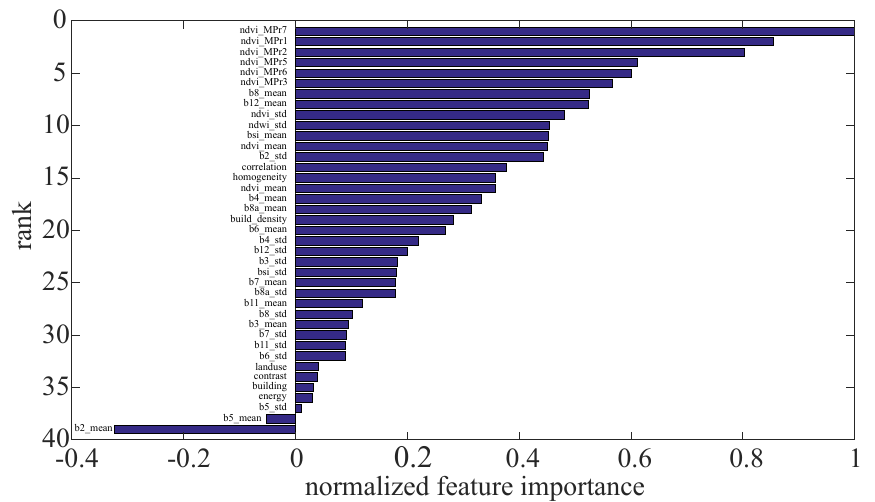}
\caption{Feature importance of the Sentinel-2 group.}
\label{fea_importance_sentinel}
\end{figure}

\begin{table}[!t]
\newcommand{\tabincell}[2]{\begin{tabular}{@{}#1@{}}#2\end{tabular}}
%% increase table row spacing, adjust to taste
\renewcommand{\arraystretch}{1.3}
% if using array.sty, it might be a good idea to tweak the value of
% \extrarowheight as needed to properly center the text within the cells
\caption{Pervious and Impervious Surface Fraction of LCZs\cite{Stewart2012}}
\label{per_impervious_surface_frac}
\centering
%% Some packages, such as MDW tools, offer better commands for making tables
%% than the plain LaTeX2e tabular which is used here.
\begin{tabular}{ccc} % |c||c|
\hline
Label & \tabincell{c}{Pervious Surface\\Fraction (\%)} & \tabincell{c}{Impervious Surface\\ Fraction (\%)}\\
\hline
Compact high-rise & 0-10 & 40-60\\
Compact mid-rise & 0-20 & 30-50\\
Compact low-rise & 0-30 & 20-50\\
Open high-rise & 30-40 & 30-40\\
Open mid-rise & 20-40 & 30-50\\
Open low-rise & 30-60 & 20-50\\
Lightweight low-rise & 0-30 & 0-20\\
Large low-rise & 0-20 & 40-50\\
Sparsely built & 60-80 & 0-20\\
Heavy industry & 40-50 & 20-40\\
Dense trees & 90-100 & 0-10 \\
Scattered trees & 90-100 & 0-10\\
Bush, scrub & 90-100 & 0-10\\
Low plants & 90-100 & 0-10\\
Bare rock or paved & 0-10 & 90-100\\
Bare soil or sand & 90-100 & 0-10\\
Water & 90-100 & 0-10\\
\hline
\end{tabular}
\end{table}

Second, both rankings of the buildings and landuse features were quite low. The landuse feature ranked the last place and the 33th place in the Landsat-8 and Sentinel-2 groups, respectively. The building feature ranked the 37th place and the 35th place in the Landsat-8 and Sentinel-2 groups, respectively. These two facts indicate that landuse and building features contribute trivial importance (even negative importance) in LCZ classification if directly using feature stacking due to the data heterogeneity. Therefore, novel models are highly necessary to fuse OSM and satellite features. 

Third, the building density feature ranked higher than the building feature itself, in 22nd and 19th place, respectively. This could be due to the building interval, one of the most distinct criteria of separating different LCZ built types, which is highly related to building density values in local areas. Building features that only delineate building areas do not contain spatial information and thus have trivial contribution to classification.
%Therefore, the building density fits better for LCZs classification than the building feature itself. \\\par

\subsection{Accuracies Improvement}
\subsubsection{Data Fusion Contest Dataset}
Fig. 9 and Fig. 10 provide the classification maps generated by the proposed framework with both fusion models and the baseline framework on four test cities. Table V compares their classification accuracies in terms of overall accuracy (OA) and kappa coefficient (kappa).

First, the OA and kappa of the proposed framework were 76.15\% and 0.72, which outperformed the baseline accuracy by 6.01\% and 7\%, respectively. Furthermore, the OA and kappa of the proposed framework still
%\sout{also} 
outperformed the winner of the 2017 IEEE GRSS Data Fusion Contest\cite{Tuia2017, Yokoya2017, Yokoya2018}
%parencite{tuia20172017,igrss_lcz_multimultimulti}
by 1.21\% and 1\%, respectively, although fewer classifiers were used in this paper.

Second, the use of the landuse or building model could significantly improve the accuracies in general. For example, compared with the baseline framework, the OA in Amsterdam increased by 11.2\% and 8.71\%, respectively.
The individual use of the landuse or building model may sometimes decrease the accuracies in some cities. For example, the OA decreased by 1.76\% after applying the landuse model to the city of Chicago and by 0.49\% after applying the building model to the city of Madrid. 
One reason could be that training samples are still quite limited to well represent the complexities of the test samples, especially when OSM data have many incomplete recordings. Another reason could be that landuse and building fusion models are applied to different areas of test cities if landuse and buildings layers lack data in different areas. Consequently, combining both models could tune the pre-classification results to the largest extent.
Although accuracies may occasionally decrease when using a single fusion model, the decreased values are much smaller than the increased values when considering the four test cities. Therefore, applying the landuse or building model solely to test cities still increases the accuracies in general. 

Third, the joint use of both fusion models always increases the accuracies significantly and is consistently better than using a single model. For example, the OA increased by 11.2\% and 8.71\%, respectively, after applying the landuse and building fusion models separately to Amsterdam, but the OA increased by 20.67\% after applying both models to Amsterdam at the same time. Additionally, the OA increased by 3.61\% and decreased by -0.49\%, respectively, after applying the landuse and building fusion models separately to Madrid, but the OA increased by 4.88\% after applying both models to Madrid at the same time. This indicates that the use of both models simultaneously is not only an improvement over using each model alone but also demonstrates how the two models can aid one another and further boost the classification performance.
%This indicates that using both models is not simply adding the improvement of single models ($3.61-0.49 \neq 4.88$). Instead, the two models could aid each other and further boost the classification performance. 

Fourth, landuse and building models impact the accuracies in different test cities to various degrees. For example, the OA increased by 20.67\% in Amsterdam, but it only increased by 0.93\% in Chicago after using both fusion models together. This may indicate that training information from only five cities is not adequate to train a generalized landuse or building fusion model, especially considering that landuse and buildings layers lack much data in certain cities, such as Xi'an, Sao Paulo, Hong Kong, etc.

\begin{figure*}[!t]
\centering
\includegraphics[width=7in]{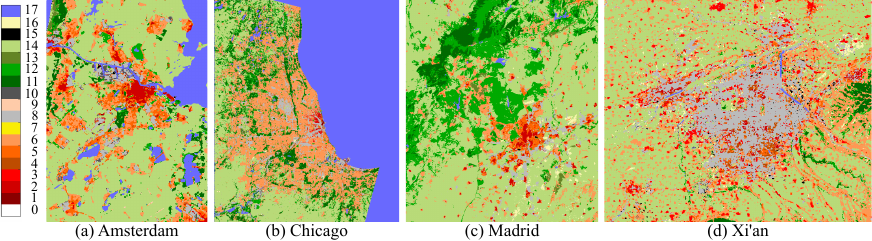}
\caption{Classification maps from proposed framework (DFC dataset).}
\label{classmap_proposed}
\end{figure*}

\begin{figure*}[!t]
\centering
\includegraphics[width=7in]{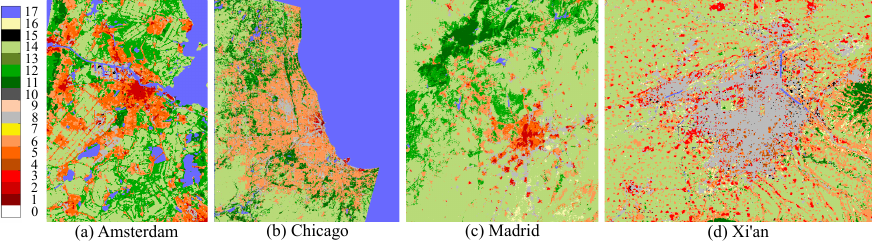}
\caption{Classification maps from baseline framework (the DFC dataset).}
\label{classmap_baseline}
\end{figure*}

\begin{table*}[!t]
\newcommand{\tabincell}[2]{\begin{tabular}{@{}#1@{}}#2\end{tabular}}
%% increase table row spacing, adjust to taste
\renewcommand{\arraystretch}{1.3}
% if using array.sty, it might be a good idea to tweak the value of
% \extrarowheight as needed to properly center the text within the cells
\caption{Classification Accuracies Comparison between the Proposed Framework and Baseline Framework (the DFC Dataset)}
\label{accu_compare_general}
\centering
%% Some packages, such as MDW tools, offer better commands for making tables
%% than the plain LaTeX2e tabular which is used here.
\begin{tabular}{ccccc} % |c||c|
\hline
City & Baseline & Landuse Fusion Model & Building Fusion Model & Both Models\\
\hline
Amsterdam & \tabincell{c}{OA=61.50\\K=0.56\\} & \tabincell{c}{OA=72.52\\K=0.68\\} & \tabincell{c}{OA=70.21\\K=0.66\\} & \tabincell{c}{OA=82.17\\K=0.79\\}\\
Chicago & \tabincell{c}{OA=77.63\\K=0.73\\} & \tabincell{c}{OA=75.87\\K=0.71\\} & \tabincell{c}{OA=78.65\\K=0.74\\} & \tabincell{c}{OA=78.56\\K=0.74\\}\\
Madrid & \tabincell{c}{OA=74.24\\K=0.69\\} & \tabincell{c}{OA=77.85\\K=0.74\\} & \tabincell{c}{OA=73.75\\K=0.69\\} & \tabincell{c}{OA=79.12\\K=0.75\\}\\
Xi'an & \tabincell{c}{OA=53.33\\K=0.44\\} & \tabincell{c}{OA=59.31\\K=0.52\\} & \tabincell{c}{OA=53.67\\K=0.45\\} & \tabincell{c}{OA=58.00\\K=0.5\\}\\
All & \tabincell{c}{OA=70.14\\K=0.65\\} & \tabincell{c}{OA=73.63\\K=0.70\\} & \tabincell{c}{OA=71.41\\K=0.67\\} & \tabincell{c}{OA=76.15\\K=0.72\\}\\
\hline
\end{tabular}
\end{table*}

Besides the OA and kappa, we also compared the improvement of the distributions of producer accuracies (PAs) (Fig. 11).
First, accuracies of all labels were improved except for label 1 (compact high-rise), where several samples were incorrectly classified as label 8 (large low-rise). The possible reasons are, first, test samples of label 1 are quite few, and the building densities between label 1 and 8 could be similar.
%This is understandable when considering the limited testing samples. 
Second, the largest improvement 
%\sout{of PA} 
occurred in label 12 (scattered trees), which was highly mixed with label 14 (low plants) in the baseline framework. 
%\sout{Due to the difference in data acquisition techniques, OSM data have more stable characteristics than satellite images when considering the spectral diversity among seasons and regions.}
A typical example exists in the classification map of Madrid (see Fig. 12). Compared with the proposed framework, the areas of scattered trees changed more significantly among different acquisition times in the case of the baseline framework. It indicates that the proposed framework is more robust to spectral changes, because OSM data have provided active aid to optical observation.
Besides, the proposed framework also demonstrates a large amount of improvement in label 10 (heavy industry), label 6 (open low-rise), label 5 (open mid-rise), label 4 (open high-rise), and label 8 (large low-rise).
%\sout{on built types}.
For example, the PA of label 10 (heavy industry) was less than 10\% through the use of the baseline framework and was highly mixed with label 8 (large low-rise), but it increased to 11\% when using the proposed framework.
Analyzing satellite images is considerably helpful for retrieving spectral, spatial, and textural information. Occasionally, retrieved knowledge of two areas from satellite images may appear to be similar (labels 8 and 10); however, these two areas belong to different LCZ labels. Compared to satellite images, OSM data contain advanced knowledge (e.g., urban functionalities) that separates different classes through human intelligence, offering non-trivial assistance in the classification of LCZs.
%In other words, it could happen that the retrieved knowledge of two areas from satellite images seems quite similar, but they belong to different LCZs classes. 

Meanwhile, the accuracies of some labels are still not satisfactory, mainly because of the complicated scheme of LCZs. 
First, label 3 (compact low-rise) and label 4 (open high-rise) are still highly mixed with other built types. It is still challenging to precisely describe urban structures due to the diverse construction of different cities.
Second, label 9 (sparsely built) is highly mixed with label 6 (open low-rise) and label 14 (low plants), likely due to the classification scale. After checking ground truth data from Google Earth, it was observed that sparsely built areas often follow a "cluster" behavior. Stated another way, in sparsely built areas, buildings appear to be dense in certain sections and absent in other sections.
%In other words, it happens in the sparsely built areas that buildings are a bit dense in part of the area while they do not appear in other parts of that area.
Third, label 15 (bare rock or paved) is highly mixed with label 8 (large low-rise), likely due to training samples. After checking ground truth data from Google Earth, it was observed that paved ground frequently appears near large low-rises and is occasionally selected for training samples of large low-rises. %usually comes out around large low-rise and sometimes are selected into the training samples of large low-rise.

\begin{figure}[!t]
\centering
\includegraphics[width=3.5in]{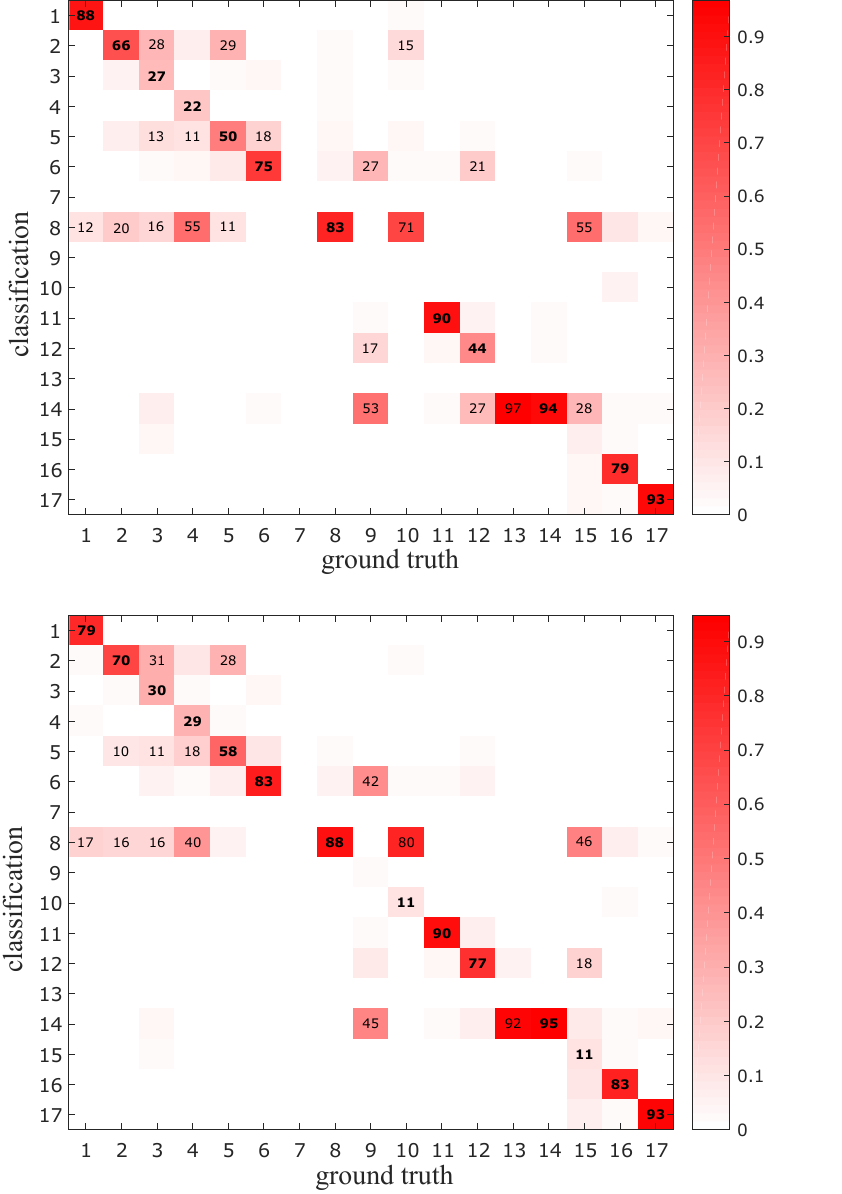}
\caption{Distributions of producer accuracies of the DFC dataset from the baseline framework (above) and the proposed framework (below). Values in the figure show the percentage of samples labeled as A in the ground truth data, which were classified as B in the classification maps, where A, B = 1,2,\ldots,17. Only the percentages above 10 were shown.}
\label{confusion_matrix_change}
\end{figure}

\begin{figure}[!t]
\centering
\includegraphics[width=3.5in]{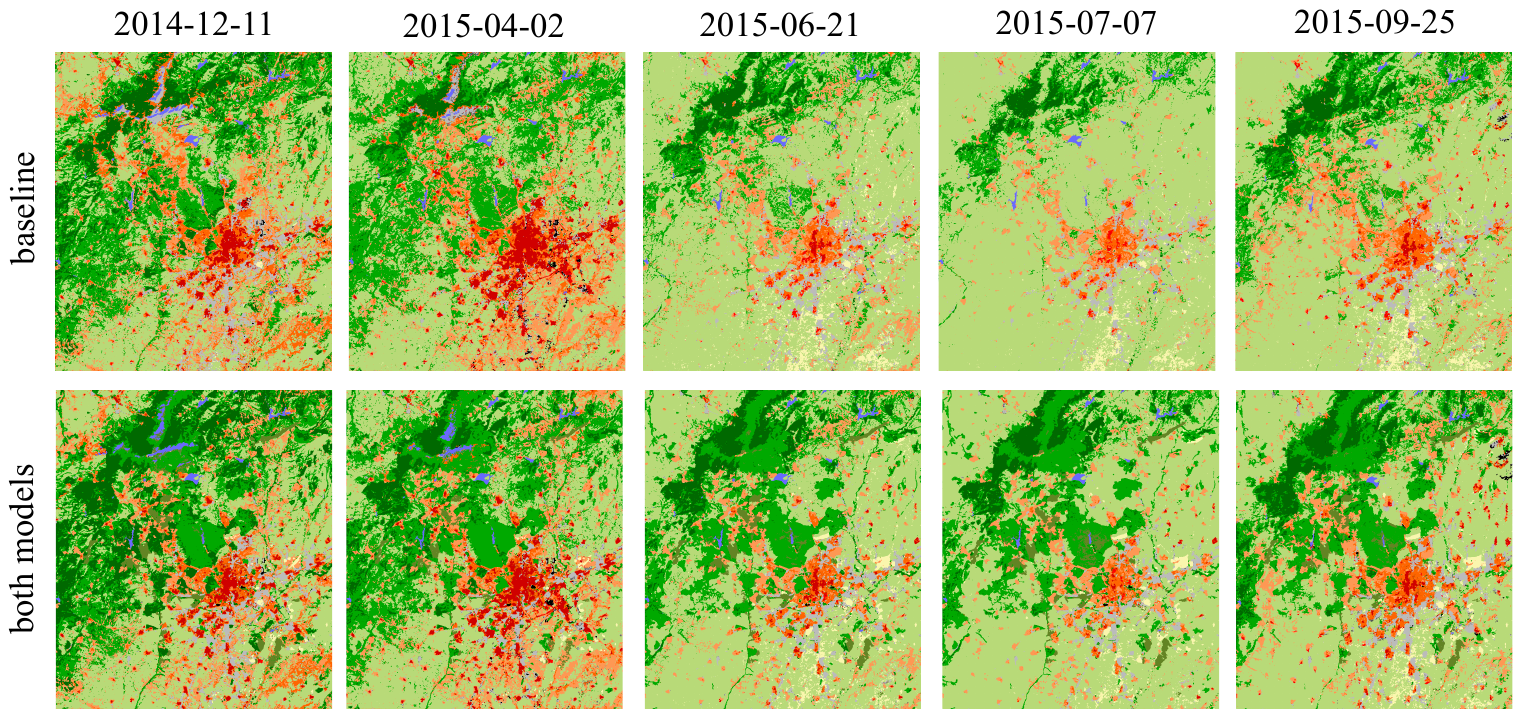}
\caption{Classification maps at each acquisition time in Madrid.}
\label{temporal_robustness_madrid}
\end{figure}

\subsubsection{Additional Test Dataset}
%First, the OA and kappa from the proposed framework in Munich were 92.51\% and 0.9, which outperform the baseline accuracy by increases by around 2\% and 4\%, respectively. The OA and kappa from the proposed framework in New York were 64.11\% and 0.57, which outperform the baseline accuracy by increases by around 2\% and 3\%, respectively.
Fig. 13 and Fig. 14 provide the classification maps generated by the proposed framework with both fusion models and the baseline framework on Munich and New York. Table VI compares their classification accuracies in terms of OA and kappa.

First, the OA and kappa from the proposed framework were 71.97\% and 0.66, which outperformed the baseline accuracy by 2.22\% and 3\%, respectively. Given the fact that the training samples were from grss\_dfc\_2017\cite{grss_dfc_2017}, the classification results demonstrated that our proposed framework was not only effective on four test cities from grss\_dfc\_2017\cite{grss_dfc_2017} but also showed satisfactory results on the newly selected test cities. Thus, the proposed framework shows advantages of transferring knowledge from the training samples in one dataset to the test samples in the other dataset based on the same classification scheme (i.e., LCZs) and data sources (i.e., satellite images and OSM data).

Second, the building fusion model improved the accuracies more significantly than the landuse fusion model. Compared with the baseline framework, the OA using the building fusion model increased by 2.12\% and 1.43\% in Munich and New York, respectively; the OA using the landuse fusion model slightly increased in Munich (0.57\%) and slightly decreased (0.26\%) in New York. Similar phenomenon also happened in the experimental results of the DFC dataset, where OA of Chicago slightly increased using the landuse fusion model and slightly decreased using the building fusion model. The joint use of both models, however, always increases the accuracies significantly and is consistently better than using a single model.

In spite of the improvement of accuracies through the proposed framework, classification accuracies of New York were notably lower than those in Munich. The possible reason could be that our proposed fusion models are based on statistical knowledge (the probability of a value in landuse/buildings layers belongs to an LCZ's label), which needs enough training samples. However, our training samples were quite limited due to the availability of a few training cities and the lack of OSM data. Moreover, OSM data may contain errors because they are open source and could be recorded by any volunteers. Another reason could be that the weighting process used in the fusion models is simple (i.e., linear) so that it may not be satisfied when dealing with complicated cases (i.e. highly non-linear).

Fig. 15 shows the improvement of the distributions of PAs for each label.
First, PAs increased on most labels, especially on label 3 (compact low-rise, 19.12\%), label 4 (open high-rise, 50.58\%), and label 12 (scattered trees, 29.01\%). Besides, PAs of label 5 (open mid-rise, 11.94\%) and label 13 (bush or scrub, 9.76\%) also increased significantly. Several samples of label 1 (compact high-rise) were misclassified as label 2 (compact mid-rise), resulting in the drop of PA on label 1. This indicated that the similarity of the building density is still a challenge to acquire satisfactory separation of all built types. Beside label 1, samples of label 9 (sparsely built) were easily misclassified to label 6 (open low-rise) and label 12 (scattered trees), likely due to the classification scale as we have mentioned in the experimental analysis of the DFC dataset.

Similar to the classification results of the DFC dataset, accuracies of some labels are still not satisfactory. The separation of different built types is still a big challenge mainly because of the complexity and diversity of urban structures and classification scale. Detailed analysis has been mentioned in IV. B. 1).
\begin{figure}[!t]
\centering
\includegraphics[width=3.5in]{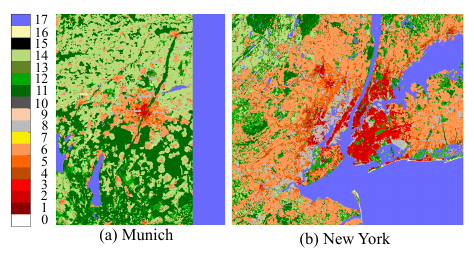}
\caption{Classification maps from proposed framework (additional test dataset).}
\label{classmap_proposed_twomorecities}
\end{figure}

\begin{figure}[!t]
\centering
\includegraphics[width=3.5in]{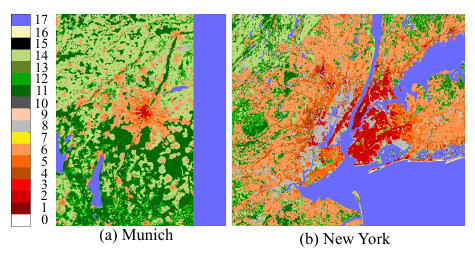}
\caption{Classification maps from baseline framework (additional test dataset).}
\label{classmap_baseline_twomorecities}
\end{figure}

\begin{table*}[!t]
\newcommand{\tabincell}[2]{\begin{tabular}{@{}#1@{}}#2\end{tabular}}
%% increase table row spacing, adjust to taste
\renewcommand{\arraystretch}{1.3}
% if using array.sty, it might be a good idea to tweak the value of
% \extrarowheight as needed to properly center the text within the cells
\caption{Classification Accuracies Comparison between the Proposed Framework and Baseline Framework (Additional Test Dataset)}
\label{accu_compare_general_twomorecities}
\centering
%% Some packages, such as MDW tools, offer better commands for making tables
%% than the plain LaTeX2e tabular which is used here.
\begin{tabular}{ccccc} % |c||c|
\hline
City & Baseline & Landuse Fusion Model & Building Fusion Model & Both Models\\
\hline
Munich & \tabincell{c}{OA=89.30\\K=0.86\\} & \tabincell{c}{OA=89.87\\K=0.87\\} & \tabincell{c}{OA=91.42\\K=0.89\\} & \tabincell{c}{OA=92.51\\K=0.90\\}\\
New York & \tabincell{c}{OA=62.26\\K=0.54\\} & \tabincell{c}{OA=62.00\\K=0.54\\} & \tabincell{c}{OA=63.69\\K=0.56\\} & \tabincell{c}{OA=64.11\\K=0.57\\}\\
All & \tabincell{c}{OA=69.75\\K=0.63\\} & \tabincell{c}{OA=69.71\\K=0.63\\} & \tabincell{c}{OA=71.37\\K=0.65\\} & \tabincell{c}{OA=71.97\\K=0.66\\}\\
\hline
\end{tabular}
\end{table*}

\begin{figure}[!t]
\centering
\includegraphics[width=3.5in]{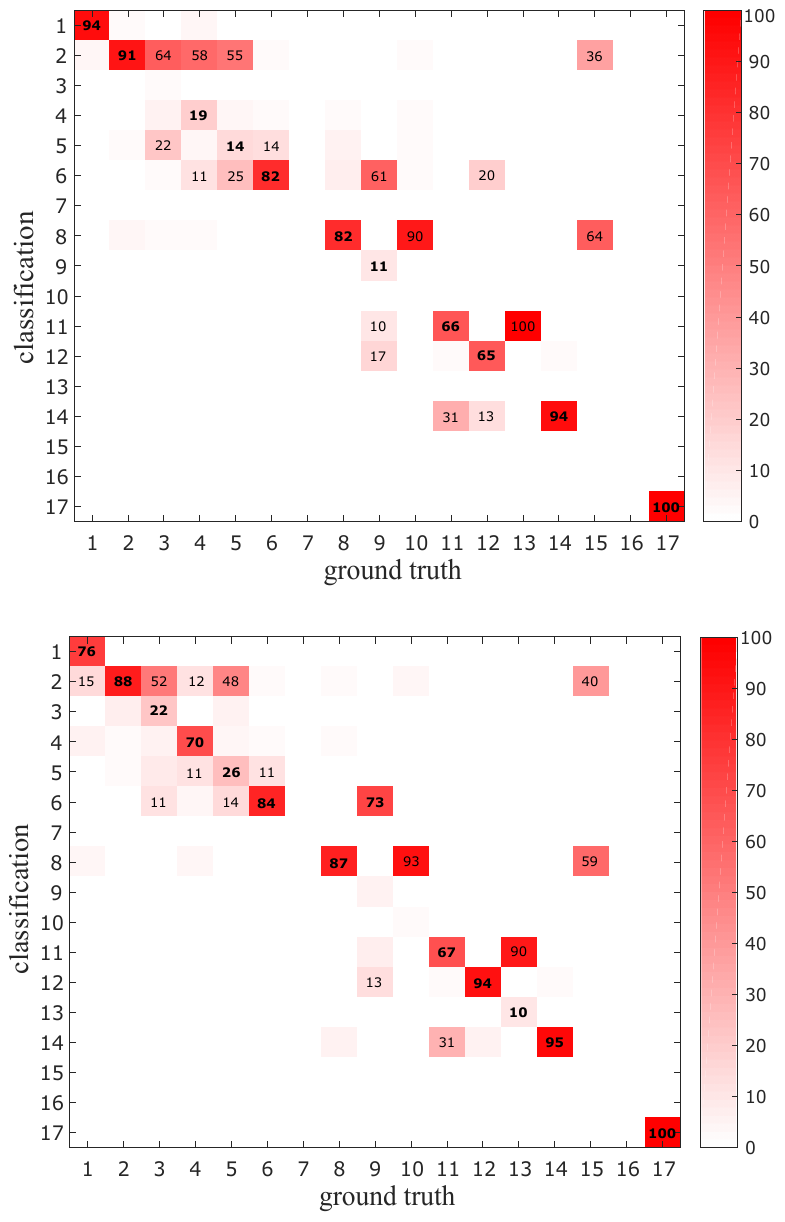}
\caption{Distributions of the producer accuracies of the additional test dataset from the baseline framework (above) and the proposed framework (below). Values in the figure show the percentage of samples labeled as A in the ground truth data, which were classified as B in the classification maps, where A, B = 1,2,\ldots,17. Only the percentages above 10 were shown.}
\label{confusion_matrix_change_additionaltestcities}
\end{figure}

\subsection{Framework Transferability}
Spectral information plays an important role in LCZ classification when using optical satellite images, but it is quite sensitive to acquisition conditions, such as time, angle, atmospheric conditions, etc. This sensitivity decreases the robustness of the classification performance, especially when considering transferability among multi-temporal images in several study areas. 

\subsubsection{Data Fusion Contest Dataset}
Tables VII, VIII, IX, X and Fig. 16 compare the classification accuracies between the proposed framework and the baseline framework using four test cities at different acquisition times. 
Results indicated that classification accuracies changed significantly among different acquisition times when applying the baseline framework to a test city. The difference in classification accuracies is still quite large, even if the acquisition times of satellite images are quite close (e.g., Amsterdam on March 12, 2015 and April 20, 2015), which indicates that this sensitivity is not only aroused by ground change but also due to other acquisition conditions. 

After applying both fusion models to test cities, this sensitivity among different acquisition times of a test city was reduced significantly. Meanwhile, classification accuracies improved at all acquisition times. 
The reason for this significant improvement is because OSM data offer positive contributions after using the proposed fusion models. 
Compared to satellite images, OSM data are not sensitive to the acquisition conditions of satellite imagery since humans can provide more advanced knowledge on recognition of the ground. This knowledge effectively tunes the classification results computed from satellite images and stabilizes the classification results aroused by spectral diversity.

\begin{figure}[!t]
\centering
\includegraphics[width=3.5in]{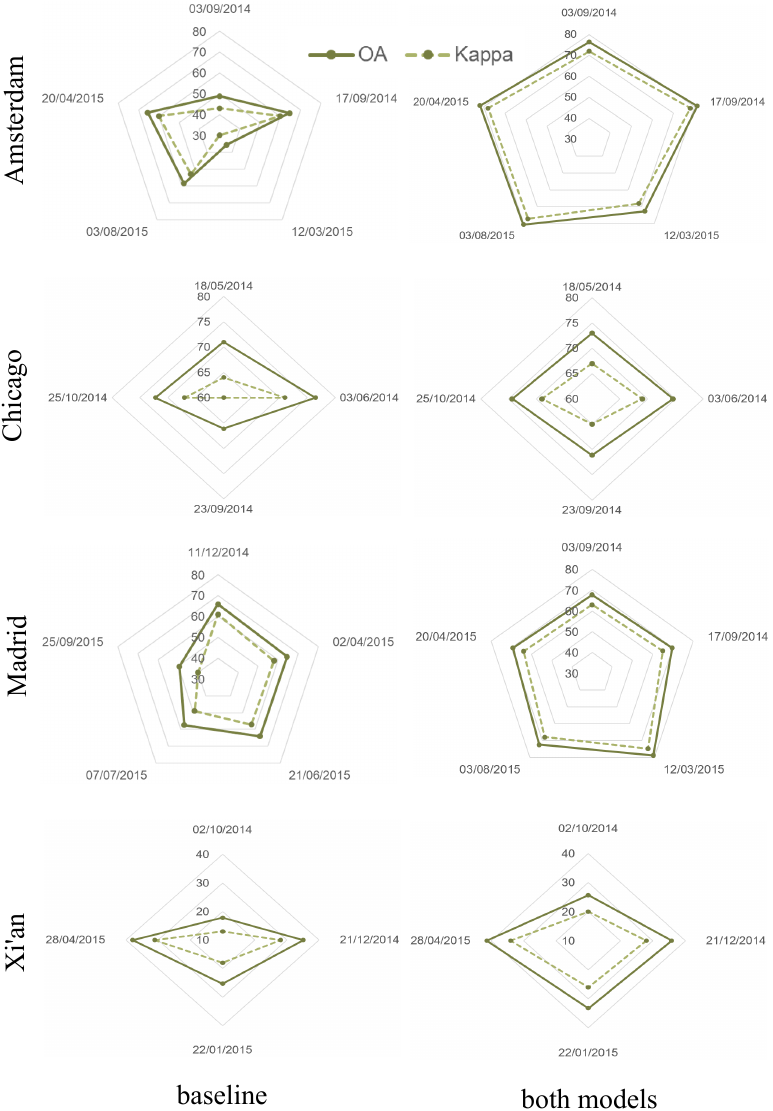}
\caption{Comparison of classification accuracies at different acquisition times (DFC dataset)}
\label{temporal_robust_dfc}
\end{figure}

%But applying only one of the fusion models may not significantly reduce this impact. 

%The reason may be the limited training samples and the transferability problem between some cities. 
% % % % Amsterdam
\begin{table*}[!t]
\newcommand{\tabincell}[2]{\begin{tabular}{@{}#1@{}}#2\end{tabular}}
%% increase table row spacing, adjust to taste
\renewcommand{\arraystretch}{1.3}
% if using array.sty, it might be a good idea to tweak the value of
% \extrarowheight as needed to properly center the text within the cells
\caption{Classification Accuracies Comparison between the Proposed Framework and Baseline Framework at Each Acquisition Time in Amsterdam}
\label{each_temporal_amsterdam}
\centering
%% Some packages, such as MDW tools, offer better commands for making tables
%% than the plain LaTeX2e tabular which is used here.
\begin{tabular}{cccccc} % |c||c|
\hline
Satellite & Date & Baseline & Landuse Model & Building Model & Both Models\\
\hline
\multirow{5}{*}{\rotatebox{90}{Landsat - 8}}& 2014-03-09 & \tabincell{c}{OA=48.90\\K=0.43\\} & \tabincell{c}{OA=67.19\\K=0.62\\} & \tabincell{c}{OA=60.43\\K=0.55\\} & \tabincell{c}{OA=76.41\\K=0.72\\}\\
& 2014-09-17 & \tabincell{c}{OA=64.57\\K=0.60\\} & \tabincell{c}{OA=74.95\\K=0.71\\} & \tabincell{c}{OA=69.50\\K=0.65\\} & \tabincell{c}{OA=81.31\\K=0.78\\}\\
& 2015-03-12 & \tabincell{c}{OA=35.60\\K=0.30\\} & \tabincell{c}{OA=61.26\\K=0.56\\} & \tabincell{c}{OA=47.62\\K=0.41\\} & \tabincell{c}{OA=72.59\\K=0.68\\}\\
 & 2015-08-03 & \tabincell{c}{OA=58.51\\K=0.53\\} & \tabincell{c}{OA=71.81\\K=0.67\\} & \tabincell{c}{OA=66.08\\K=0.61\\} & \tabincell{c}{OA=80.46\\K=0.77\\}\\
 & 2015-04-20 & \tabincell{c}{OA=65.52\\K=0.60\\} & \tabincell{c}{OA=77.21\\K=0.73\\} & \tabincell{c}{OA=68.21\\K=0.63\\} & \tabincell{c}{OA=81.88\\K=0.78\\}\\
\hline
Sentinel - 2 & 2016-09-08 & \tabincell{c}{OA=59.76\\K=0.54\\} & \tabincell{c}{OA=70.91\\K=0.66\\} & \tabincell{c}{OA=70.34\\K=0.66\\} & \tabincell{c}{OA=81.98\\K=0.78\\}\\
\hline
\end{tabular}
\end{table*}

% % % % Chicago
\begin{table*}[!t]
\newcommand{\tabincell}[2]{\begin{tabular}{@{}#1@{}}#2\end{tabular}}
%% increase table row spacing, adjust to taste
\renewcommand{\arraystretch}{1.3}
% if using array.sty, it might be a good idea to tweak the value of
% \extrarowheight as needed to properly center the text within the cells
\caption{Classification Accuracies Comparison between the Proposed Framework and Baseline Framework at Each Acquisition Time in Chicago}
\label{each_temporal_chicago}
\centering
%% Some packages, such as MDW tools, offer better commands for making tables
%% than the plain LaTeX2e tabular which is used here.
\begin{tabular}{cccccc} % |c||c|
\hline
Satellite & Date & Baseline & Landuse Model & Building Model & Both Models\\
\hline
\multirow{5}{*}{\rotatebox{90}{Landsat - 8}}& 2014-05-18 & \tabincell{c}{OA=71.01\\K=0.64\\} & \tabincell{c}{OA=71.07\\K=0.65\\} & \tabincell{c}{OA=70.85\\K=0.64\\} & \tabincell{c}{OA=72.97\\K=0.67\\}\\
& 2014-06-03 & \tabincell{c}{OA=76.46\\K=0.71\\} & \tabincell{c}{OA=72.31\\K=0.66\\} & \tabincell{c}{OA=75.11\\K=0.70\\} & \tabincell{c}{OA=74.56\\K=0.69\\}\\
& 2014-09-23 & \tabincell{c}{OA=66.09\\K=0.6\\} & \tabincell{c}{OA=66.78\\K=0.66\\} & \tabincell{c}{OA=70.10\\K=0.64\\} & \tabincell{c}{OA=71.11\\K=0.65\\}\\
 & 2014-10-25 & \tabincell{c}{OA=72.21\\K=0.67\\} & \tabincell{c}{OA=69.42\\K=0.63\\} & \tabincell{c}{OA=73.67\\K=0.68\\} & \tabincell{c}{OA=74.31\\K=0.69\\}\\
\hline
Sentinel - 2 & 2016-10-13 & \tabincell{c}{OA=55.25\\K=0.49\\} & \tabincell{c}{OA=55.61\\K=0.49\\} & \tabincell{c}{OA=69.57\\K=0.64\\} & \tabincell{c}{OA=69.63\\K=0.64\\}\\
\hline
\end{tabular}
\end{table*}

% Madrid
\begin{table*}[!t]
\newcommand{\tabincell}[2]{\begin{tabular}{@{}#1@{}}#2\end{tabular}}
%% increase table row spacing, adjust to taste
\renewcommand{\arraystretch}{1.3}
% if using array.sty, it might be a good idea to tweak the value of
% \extrarowheight as needed to properly center the text within the cells
\caption{Classification Accuracies Comparison between the Proposed Framework and Baseline Framework at Each Acquisition Time in Madrid}
\label{each_temporal_madrid}
\centering
%% Some packages, such as MDW tools, offer better commands for making tables
%% than the plain LaTeX2e tabular which is used here.
\begin{tabular}{cccccc} % |c||c|
\hline
Satellite & Date & Baseline & Landuse Model & Building Model & Both Models\\
\hline
\multirow{5}{*}{\rotatebox{90}{Landsat - 8}}& 2014-12-11 & \tabincell{c}{OA=65.92\\K=0.61\\} & \tabincell{c}{OA=67.80\\K=0.63\\} & \tabincell{c}{OA=65.14\\K=0.60\\} & \tabincell{c}{OA=67.80\\K=0.63\\}\\
& 2015-04-02 & \tabincell{c}{OA=64.44\\K=0.58\\} & \tabincell{c}{OA=68.85\\K=0.64\\} & \tabincell{c}{OA=62.69\\K=0.57\\} & \tabincell{c}{OA=69.70\\K=0.65\\}\\
& 2015-06-21 & \tabincell{c}{OA=63.94\\K=0.57\\} & \tabincell{c}{OA=78.07\\K=0.74\\} & \tabincell{c}{OA=64.50\\K=0.58\\} & \tabincell{c}{OA=78.94\\K=0.75\\}\\
 & 2015-07-07 & \tabincell{c}{OA=57.35\\K=0.49\\} & \tabincell{c}{OA=69.73\\K=0.64\\} & \tabincell{c}{OA=57.61\\K=0.49\\} & \tabincell{c}{OA=72.52\\K=0.68\\}\\
 & 2015-09-25 & \tabincell{c}{OA=49.38\\K=0.40\\} & \tabincell{c}{OA=67.48\\K=0.62\\} & \tabincell{c}{OA=51.23\\K=0.42\\} & \tabincell{c}{OA=69.23\\K=0.64\\}\\
\hline
Sentinel - 2 & 2016-10-11 & \tabincell{c}{OA=68.96\\K=0.63\\} & \tabincell{c}{OA=73.72\\K=0.69\\} & \tabincell{c}{OA=67.21\\K=0.61\\} & \tabincell{c}{OA=74.92\\K=0.71\\}\\
\hline
\end{tabular}
\end{table*}

% % % Xi'an
\begin{table*}[!t]
\newcommand{\tabincell}[2]{\begin{tabular}{@{}#1@{}}#2\end{tabular}}
%% increase table row spacing, adjust to taste
\renewcommand{\arraystretch}{1.3}
% if using array.sty, it might be a good idea to tweak the value of
% \extrarowheight as needed to properly center the text within the cells
\caption{Classification Accuracies Comparison between the Proposed Framework and Baseline Framework at Each Acquisition Time in Xi'an}
\label{each_temporal_xian}
\centering
%% Some packages, such as MDW tools, offer better commands for making tables
%% than the plain LaTeX2e tabular which is used here.
\begin{tabular}{cccccc} % |c||c|
\hline
Satellite & Date & Baseline & Landuse Model & Building Model & Both Models\\
\hline
\multirow{5}{*}{\rotatebox{90}{Landsat - 8}}& 2014-10-02 & \tabincell{c}{OA=17.78\\K=0.13\\} & \tabincell{c}{OA=25.15\\K=0.19\\} & \tabincell{c}{OA=16.92\\K=0.12\\} & \tabincell{c}{OA=25.72\\K=0.20\\}\\
& 2014-12-21 & \tabincell{c}{OA=35.05\\K=0.28\\} & \tabincell{c}{OA=35.41\\K=0.28\\} & \tabincell{c}{OA=31.26\\K=0.24\\} & \tabincell{c}{OA=35.81\\K=0.28\\}\\
& 2015-01-22 & \tabincell{c}{OA=25.28\\K=0.18\\} & \tabincell{c}{OA=32.48\\K=0.25\\} & \tabincell{c}{OA=23.35\\K=0.16\\} & \tabincell{c}{OA=33.22\\K=0.26\\}\\
 & 2015-04-28 & \tabincell{c}{OA=37.88\\K=0.31\\} & \tabincell{c}{OA=41.27\\K=0.34\\} & \tabincell{c}{OA=36.49\\K=0.29\\} & \tabincell{c}{OA=41.35\\K=0.34\\}\\
\hline
Sentinel - 2 & 2016-08-27 & \tabincell{c}{OA=53.33\\K=0.44\\} & \tabincell{c}{OA=59.31\\K=0.52\\} & \tabincell{c}{OA=53.67\\K=0.45\\} & \tabincell{c}{OA=58.00\\K=0.50\\}\\
\hline
\end{tabular}
\end{table*}

Furthermore, the increased transferability of the proposed framework allows higher classification performance with less temporal information. Table XI compares the classification accuracies after applying frameworks to the multi-temporal Landsat-8 group\footnote{Multi-temporal Landsat-8 group contains multi-temporal Landsat-8 images and OSM data.} and single-temporal Sentinel-2 group\footnote{Single-temporal Sentinel-2 group contains single-temporal Sentinel-2 images and OSM data.} of test cities. When applying frameworks to multi-temporal data, frameworks conduct majority voting among different acquisition times of each test city. 

After applying the baseline framework to all test cities, the OA difference between the Landsat-8 and Sentinel-2 groups is about 5.61\%. However, this difference drops to around 0.84\% after applying the proposed framework to all test cities. The city of Xi'an contributes greatly here as the OA of Xi'an after applying the Sentinel-2 group is much higher than the OA after applying 
%\textcolor{red}{applying what? when you use the word "apply" the sentence should be like applying X to Y]}
the Landsat-8 group, which is a quite interesting phenomenon for further studies. 
Moreover, we compare the OA difference in three other cities in order to remove the impact of Xi'an. After applying the baseline framework to Amsterdam, Chicago, and Xi'an, the OA difference of the three test cities, between the Landsat-8 and Sentinel-2 groups, is about 9.1\%. However, this difference drops to around 3.94\% after applying the proposed framework to these three test cities.
These results indicate that the proposed framework could significantly improve classification performance and acquire more trustworthy classification results with less temporal information. 
This advantage is quite useful for urban classification when ground change should be prevented. Multi-temporal data may boost classification performance by taking temporal-spectral variability into consideration\cite{Yokoya2017}. Meanwhile, when applying frameworks to multi-temporal data, the ground truth may not correspond to all data due to ground change along different acquisition times, which becomes a considerable problem. 
The proposed framework significantly shrinks the gap between using multi-temporal and single-temporal data. Therefore, single-temporal data could also achieve satisfactory classification performance if multi-temporal data are not available or they contain many ground changes.
%When applying the baseline framework to all testing cities, the OA of Landsat-8 group and Sentinel-2 group is 67.08 \% and 61.47 \% respectively. When applying the proposed framework with both fusion models to all testing cities, the OA of Landsat-8 group and Sentinel-2 group becomes 72.51 \% and 71.67 \% respectively. These results indicate that 
%\textcolor{red}{These results indicate that the proposed framework could significantly boost classification performance, meanwhile acquire a more trustworthy classification results with limited temporal information.}  \\\par

Although the proposed framework could reduce the gap between OAs from the Landsat-8 group and Sentinel-2 group, classification performance is still stronger when using multi-temporal, rather than single-temporal data for the following reasons. First, multi-temporal data provides more opportunities to record the ground reflectance in several data acquisition times so that it improves the transferability between cities by increasing the spectral diversity of training and test cities. For example, the spectral reflectance of the same ground object may change occasionally due to various data acquisition conditions, thus single temporal images may not fully represent the spectral information of ground objects.
Second, multi-temporal images could alleviate the cloud impact. The areas covered by clouds contain limited ground information, but it can be assumed that the clouds appear in different areas at different acquisition times. The majority voting among several classification maps generated from multi-temporal images could mostly remove the cloud impact. For example, the OA from Sentinel-2 in Chicago is significantly lower than the OA from Landsat-8, likely because the Sentinel-2 images in Chicago have high cloud coverage. 

% % % 
\begin{table*}[!t]
\newcommand{\tabincell}[2]{\begin{tabular}{@{}#1@{}}#2\end{tabular}}
%% increase table row spacing, adjust to taste
\renewcommand{\arraystretch}{1.3}
% if using array.sty, it might be a good idea to tweak the value of
% \extrarowheight as needed to properly center the text within the cells
\caption{Classification Accuracies Comparison between Landsat-8 and Sentinel-2}
\label{temporal_landsat_sentinel}
\centering
%% Some packages, such as MDW tools, offer better commands for making tables
%% than the plain LaTeX2e tabular which is used here.
\begin{tabular}{cccccc} % |c||c|
\hline
Satellite & City & Baseline & Landuse Model & Building Model & Both Models\\
\hline
\multirow{5}{*}{\rotatebox{90}{Landsat - 8}}& Amsterdam & \tabincell{c}{OA=59.74\\K=0.55\\} & \tabincell{c}{OA=72.34\\K=0.68\\} & \tabincell{c}{OA=68.08\\K=0.63\\} & \tabincell{c}{OA=81.12\\K=0.77\\}\\
& Chicago & \tabincell{c}{OA=78.15\\K=0.73\\} & \tabincell{c}{OA=76.43\\K=0.71\\} & \tabincell{c}{OA=78.80\\K=0.74\\} & \tabincell{c}{OA=78.56\\K=0.74\\}\\
& Madrid & \tabincell{c}{OA=73.37\\K=0.68\\} & \tabincell{c}{OA=78.18\\K=0.74\\} & \tabincell{c}{OA=72.71\\K=0.67\\} & \tabincell{c}{OA=78.67\\K=0.75\\}\\
 & Xi'an & \tabincell{c}{OA=37.16\\K=0.31\\} & \tabincell{c}{OA=37.74\\K=0.31\\} & \tabincell{c}{OA=31.61\\K=0.25\\} & \tabincell{c}{OA=38.48\\K=0.31\\}\\
 & All & \tabincell{c}{OA=67.08\\K=0.62\\} & \tabincell{c}{OA=70.44\\K=0.66\\} & \tabincell{c}{OA=67.17\\K=0.62\\} & \tabincell{c}{OA=72.51\\K=0.68\\}\\
\hline
\multirow{5}{*}{\rotatebox{90}{Sentinel - 2}} & Amsterdam & \tabincell{c}{OA=59.76\\K=0.54\\} & \tabincell{c}{OA=70.91\\K=0.66\\} & \tabincell{c}{OA=70.34\\K=0.66\\} & \tabincell{c}{OA=81.98\\K=0.78\\}\\
 & Chicago & \tabincell{c}{OA=55.25\\K=0.49\\} & \tabincell{c}{OA=55.61\\K=0.49\\} & \tabincell{c}{OA=69.57\\K=0.64\\} & \tabincell{c}{OA=69.63\\K=0.64\\}\\
 & Madrid & \tabincell{c}{OA=68.96\\K=0.63\\} & \tabincell{c}{OA=73.72\\K=0.69\\} & \tabincell{c}{OA=67.21\\K=0.61\\} & \tabincell{c}{OA=74.92\\K=0.71\\}\\
 & Xi'an & \tabincell{c}{OA=53.33\\K=0.44\\} & \tabincell{c}{OA=59.31\\K=0.52\\} & \tabincell{c}{OA=53.67\\K=0.45\\} & \tabincell{c}{OA=58.00\\K=0.50\\}\\
 & All & \tabincell{c}{OA=61.47\\K=0.56\\} & \tabincell{c}{OA=66.05\\K=0.61\\} & \tabincell{c}{OA=66.10\\K=0.61\\} & \tabincell{c}{OA=71.67\\K=0.67\\}\\
\hline
\end{tabular}
\end{table*}

\subsubsection{Addition Test Dataset}
Tables XII, XIII, and Fig. 17 compare the classification accuracies of the proposed framework and the baseline framework using two test cities at different acquisition times. Similar to the results on the DFC dataset, classification accuracies of the baseline framework on the additional test cities were sensitive to the acquisition times of the satellite images. For example, the kappa of Munich in April and July 2017 was 0.84 and it decreased to 0.78 in October 2017. Kappa values of New York in June and October 2017 were 0.56 and 0.54, respectively, and the kappa value of New York in April 2018 dropped to 0.49.

After applying the proposed framework to test cities, the classification results were less sensitive to different acquisition times. Meanwhile, classification accuracies improved at all acquisition times. For example, kappa values of Munich at three acquisition times ranged from 0.89 to 0.90, and the kappa values of New York ranged from 0.53 to 0.59. 
Therefore, it can be concluded that the proposed framework also demonstrated satisfactory transferability on the additional test dataset. The detailed analysis of the framework transferability can be found in IV. C. 1).

%%% munich
\begin{table*}[!t]
\newcommand{\tabincell}[2]{\begin{tabular}{@{}#1@{}}#2\end{tabular}}
%% increase table row spacing, adjust to taste
\renewcommand{\arraystretch}{1.3}
% if using array.sty, it might be a good idea to tweak the value of
% \extrarowheight as needed to properly center the text within the cells
\caption{Classification Accuracies Comparison between the Proposed Framework and Baseline Framework at Each Acquisition Time in Munich}
\label{each_temporal_munich}
\centering
%% Some packages, such as MDW tools, offer better commands for making tables
%% than the plain LaTeX2e tabular which is used here.
\begin{tabular}{cccccc} % |c||c|
\hline
Satellite & Date & Baseline & Landuse Model & Building Model & Both Models\\
\hline
\multirow{5}{*}{\rotatebox{90}{Sentinel-2}}& 2017-04-24 & \tabincell{c}{OA=87.71\\K=0.84\\} & \tabincell{c}{OA=89.91\\K=0.87\\} & \tabincell{c}{OA=88.00\\K=0.85\\} & \tabincell{c}{OA=92.29\\K=0.90\\}\\
& 2017-07-18 & \tabincell{c}{OA=87.75\\K=0.84\\} & \tabincell{c}{OA=89.02\\K=0.86\\} & \tabincell{c}{OA=89.57\\K=0.87\\} & \tabincell{c}{OA=91.24\\K=0.89\\}\\
& 2017-10-16 & \tabincell{c}{OA=82.24\\K=0.78\\} & \tabincell{c}{OA=90.05\\K=0.87\\} & \tabincell{c}{OA=85.50\\K=0.82\\} & \tabincell{c}{OA=91.79\\K=0.90\\}\\
\hline
\end{tabular}
\end{table*}

%%% new york
\begin{table*}[!t]
\newcommand{\tabincell}[2]{\begin{tabular}{@{}#1@{}}#2\end{tabular}}
%% increase table row spacing, adjust to taste
\renewcommand{\arraystretch}{1.3}
% if using array.sty, it might be a good idea to tweak the value of
% \extrarowheight as needed to properly center the text within the cells
\caption{Classification Accuracies Comparison between the Proposed Framework and Baseline Framework at Each Acquisition Time in New York}
\label{each_temporal_newyork}
\centering
%% Some packages, such as MDW tools, offer better commands for making tables
%% than the plain LaTeX2e tabular which is used here.
\begin{tabular}{cccccc} % |c||c|
\hline
Satellite & Date & Baseline & Landuse Model & Building Model & Both Models\\
\hline
\multirow{5}{*}{\rotatebox{90}{Sentinel-2}}& 2017-06-12 & \tabincell{c}{OA=63.78\\K=0.56\\} & \tabincell{c}{OA=63.38\\K=0.56\\} & \tabincell{c}{OA=67.86\\K=0.61\\} & \tabincell{c}{OA=66.25\\K=0.59\\}\\
& 2017-10-20 & \tabincell{c}{OA=61.45\\K=0.54\\} & \tabincell{c}{OA=63.83\\K=0.57\\} & \tabincell{c}{OA=63.67\\K=0.56\\} & \tabincell{c}{OA=64.05\\K=0.57\\}\\
& 2018-04-23 & \tabincell{c}{OA=57.83\\K=0.49\\} & \tabincell{c}{OA=59.59\\K=0.51\\} & \tabincell{c}{OA=60.93\\K=0.53\\} & \tabincell{c}{OA=61.60\\K=0.53\\}\\
\hline
\end{tabular}
\end{table*}

\begin{figure}[!t]
\centering
\includegraphics[width=3.5in]{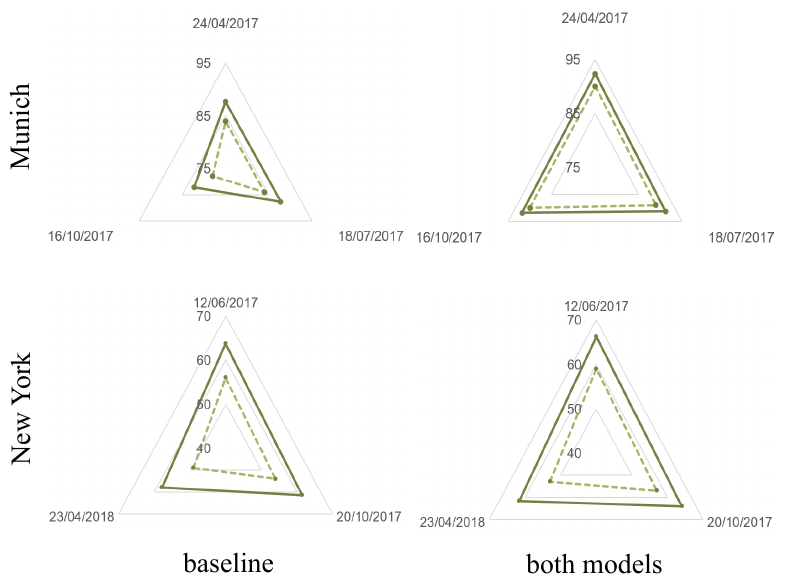}
\caption{Comparison of classification accuracies at different acquisition times (additional test dataset)}
\label{temporal_robust_additionaltest}
\end{figure}

\subsection{Building Confidence Masks}
Building confidence masks remove most incomplete building areas and provide much cleaner building density features for generating the building weight matrix. In this section, we generate building weight matrices using different building confidence masks with various approaches and thresholds (Fig. 19). Fig. 19 (a) to (k) present the matrices after applying the proposed approach to generating building confidence masks (the thresholds of the building surface fraction range from 0\% to 100\% with the step of 10\%). Fig. 19 (l) indicates the matrix after using the rule of the building surface fraction (Table III) to generate building confidence masks. Fig. 19 (m) demonstrates the matrix after using all-pass masks.
All matrices were computed by using training samples as computing the matrix (l) in Fig. 19 required ground truth data.
The matrix (l) in Fig. 19 could be regarded as quasi-truth. 
It is assumed that an effective approach to generating building confidence masks should have a high correlation with the matrix (l) in Fig. 19 and a low correlation with the matrix (m) in Fig. 19.
The correlation between the matrices generated from different building confidence masks are investigated in Fig. 20. Fig. 20 (a) presents the correlation between the matrices of Fig. 19 (a) -- (k) and the matrix of Fig. 19 (l) and (b) illustrates the correlation between the matrices of Fig. 19 (a) -- (k) and the matrix of Fig. 19 (m).

The results indicate that, firstly, building confidence masks are not sensitive to the threshold of a building surface fraction between 10 and 70. Second, the correlation significantly increased from a threshold of 0 to 10 and then remained stable between 10 and 70, indicating that a threshold of 10 is acceptable for this paper. Third, the correlation became unsatisfied after a threshold of 70 because the high threshold masked out many areas that should have been retained. Fourth, these two correlation graphs not only indicate that the generation of building confidence masks is not sensitive to the threshold of a building surface fraction but also provide a way to automatically select the threshold. For example, a threshold of 50 should be selected to retain a larger correlation in Fig. 20 (a) and a lower correlation in Fig. 20 (b). 
We set the threshold as 10 for the following two reasons.
First, this parameter is not sensitive. Second, sparsely built areas can be retained, because 10 \% is the lowest boundary of the building surface fraction of that label.
%we would like to keep the sparsely built because 10 is the lower boundary of the building surface fraction of that label. 

Although this approach is effective from a statistical point of view, several open issues remain. Certain empirical parameters are included, such as the searching radius, threshold of the building surface fraction, and threshold of generating masks. 
Since very high-resolution images are not available in the dataset\cite{grss_dfc_2017}, this approach still initiates a novel idea about building data validation by considering the relations among OSM data. 

\begin{figure}[!t]
\centering
\includegraphics[width=3.5in]{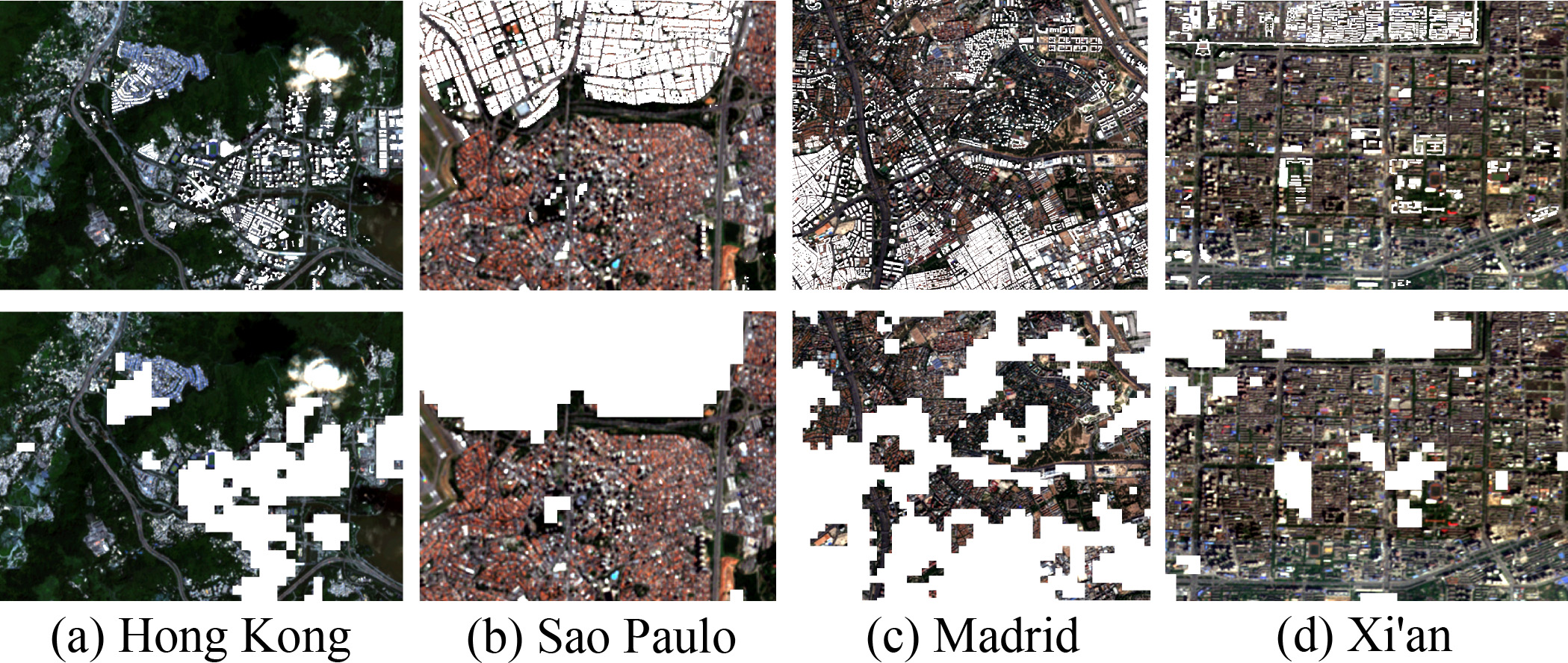}
\caption{Examples of building confidence masks (upper: a buildings layer overlaid with Sentinel-2 images; below: a building confidence mask overlaid with Sentinel-2 images).}
\label{example_build_confmask}
\end{figure}

% statistics
\begin{figure}[!t]
\centering
\includegraphics[width=3.5in]{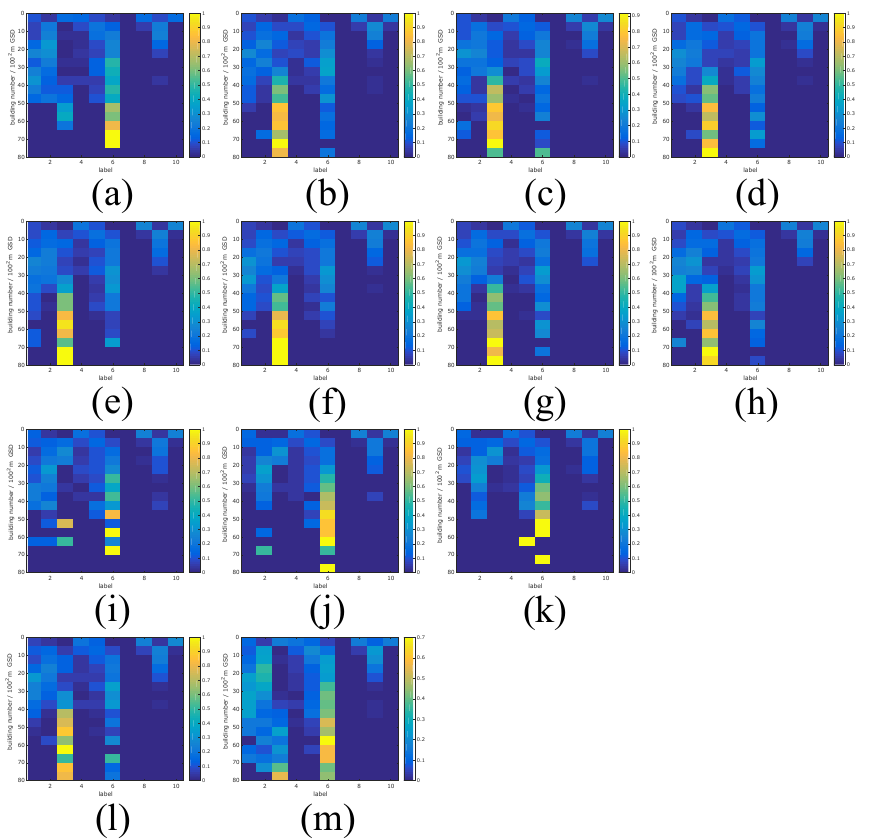}
\caption{A comparison of building densities -- label distributions among different approaches to generating building confidence masks under various thresholds of building surface fraction. (a)-(k) proposed approach using the building surface fraction thresholds from 0\% to 100\% with a step of 10\%; (l) building surface fraction rule; (m) all-pass masks.}
\label{statis_compare_buildconf}
\end{figure}

\begin{figure}[!t]
\centering
\includegraphics[width=3.5in]{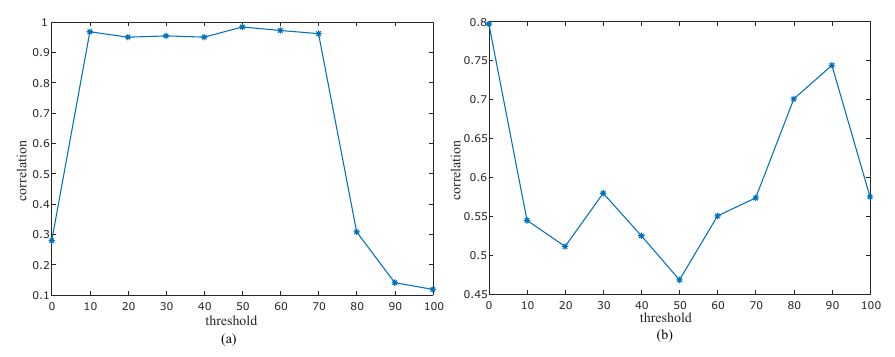}
\caption{Correlation between different building density - label distributions. (a) between Fig. 19 (a) - (k) and Fig. 20 (l); (b) between Fig. 19 (a) - (k) and Fig. 20 (m).}
\label{correlation_compare_buildconf}
\end{figure}

%\begin{figure}[!t]
%\centering
%\includegraphics[width=3.5in]{figures/buildingdensity_label_3rulecompare.pdf}
%\caption{Building Density - Label Distribution comparison among different methods of generating building confidence masks. (a) the proposed building confidence masks; (b) without using building confidence masks; (c) using the building surface fraction standard}
%\label{statis_compare_buildconf}
%\end{figure}

\section{Discussion}
The proposed framework with two fusion models acquires higher classification accuracies while additionally achieving more stable and generalized results on a worldwide scale. 
First, feature stacking is the most commonly used and fastest implemented approach of feature fusion. It stacks features along the feature dimension and with the intent that the classifier(s) will find the satisfactory hyperplane, but this is not always the case. When features are heterogeneous, such as having different noise sources, it will add many outliers in the feature space, which significantly impacts hyperplane splits. This paper has resolved this heterogeneity problem by embedding the relation among multi-source data in the proposed fusion models. This presents a novel idea to fuse different data sources by considering both features and their deep relations. 

% After all, the fusion problem is not only to add the information of multi-sources, but also to add the information of multi-sources and their deep relations. \\\par
Second, it is of great interest to understand how the data sources achieve optimal collaboration.
% the contributions that each data source could play the key to the success of achieving more stable classification performance is that the proposed framework properly considers which role that OSM should play in the classification.
OSM data containing human intelligence should have more advanced and stable knowledge of the ground objects, compared with satellite images, but they contain incomplete and incorrect recordings. Satellite images recorded by sensors typically have more objective and abundant information from the ground; however, they are quite sensitive to acquisition conditions. In order to stabilize the classification performance, the proposed framework firstly creates an initial classification result through satellite images and then tunes the initial result by using OSM data. 

%Considering their advantages and disadvantages, the proposed framework firstly generates an initial classification result using satellite images, then removes the noisy pixels of OSM and tunes the initial result through OSM. \\\par

%Meanwhile, LCZs classification are still facing many challenges. 
The proposed framework still contains several open problems. 
First, the proposed fusion models use a simple 2D probability distribution to map the relation between OSM and satellite features, which may be not accurate. Therefore, more complicated models that can embed deeper relations, such as association rule learning\cite{Agrawal1993}, can be derived. 
Second, to avoid a negative impact, we have resolved the problems of the data incompleteness of OSM, but we omitted the problem of incorrect recordings from OSM data.
%we have managed to deal with data incompleteness of OSM while we omit the problem of incorrect recordings which could also cause negative impact.
Third, certain empirical parameters are needed to generate building confidence masks.
Fourth, because of multi-source and multi-temporal data, the acquisition times of all data sources are not consistent, which could lead to many problems in analyzing the data without discrimination.
Fourth, the model transferability between cities requires further research. Due to the high cost of selecting the training samples, it would be ideal to transfer the knowledge trained from some cities to classify other cities. CCF has already successfully demonstrated how to resolve this problem, by embedding the correlation between features and labels in the projected feature space. However, this solution may not be accurate or the best method.
Lastly, urban scenes contain many intermediate LCZ labels. Each label of LCZs is defined as a certain combination of several ground types with certain structure arrangements that have only a few range values, instead of quantitatively precise definitions, to describe each label\cite{Stewart2012}. This could generate many intermediate areas, which may belong to multi-labels.%~\parencite{stewart2012local}

\section{Conclusion}
This paper proposes a framework with two novel models of fusing satellite optical images and OSM data for the classification of LCZs.
The contributions of extracted features have been investigated, and it has been discovered that OSM features possess trivial or even negative contributions to the classification performance through the use of feature stacking.
Accordingly, we proposed a new framework that embeds the multi-source data and their relations by considering the data heterogeneity of 
%\sout{satellite} 
optical images and OSM data. The proposed framework achieves better classification performance than state-of-the-art frameworks. Furthermore, its increased robustness shows promise in generalizing the framework for a worldwide scale with less temporal information use.
%Meanwhile, it has more robust behavior so that it is more promising to generalize the framework on a worldwide scale with less temporal information using. 
%Compared to the novel fusion models,  when applying feature stacking.
In addition, this paper introduces the novel idea of detecting incomplete data areas in buildings layers by considering their relation with landuse layers. This approach offers much assistance in building incompleteness detection when very high-resolution images are not available. 

%Overall, this paper proposes a novel fusion idea of dealing with heterogeneous data by considering their relations. 
This paper contains certain open issues that are of interest for further investigation. First, there is the prospect of forming more sophisticated fusion models that embed additional complex rules between heterogeneous data. Specifically, it can be an excellent future work to extend this framework to fuse other data sources, such as SAR images. Second, solving the problem of incomplete and inaccurate recordings in OSM data remains a challenging topic.
Third, there is a need for the investigation of knowledge transferability among different study regions in order to generate classification maps on a global scale.

%\appendices

%\section{}

\section*{Acknowledgment}
The authors would like to thank the WUDAPT( http://www.wudapt.org/ ) and GeoWIKI ( http://geo-wiki.org/ ) initiatives for providing the data packages used in this study, the DASE benchmarking platform ( http://dase.ticinumaerospace.com/ ), and the IEEE GRSS Image Analysis and Data Fusion Technical Committee. Landsat 8 data are available from the U.S. Geological Survey (https://www.usgs.gov/). OpenStreetMap Data © OpenStreetMap contributors, available under the Open Database Licence – http://www.openstreetmap.org/copyright. Original Copernicus Sentinel Data 2016 are available from the European Space Agency (https://sentinel.esa.int).

\ifCLASSOPTIONcaptionsoff
  \newpage
\fi



\bibliographystyle{IEEEtran}
\bibliography{IEEEabrv,bibtex/bib/bibiliograhpy}

%\begin{thebibliography}{1}


%\end{thebibliography}


 \begin{IEEEbiography}
 [{\includegraphics[width=1in,height=1.25in,clip,keepaspectratio]{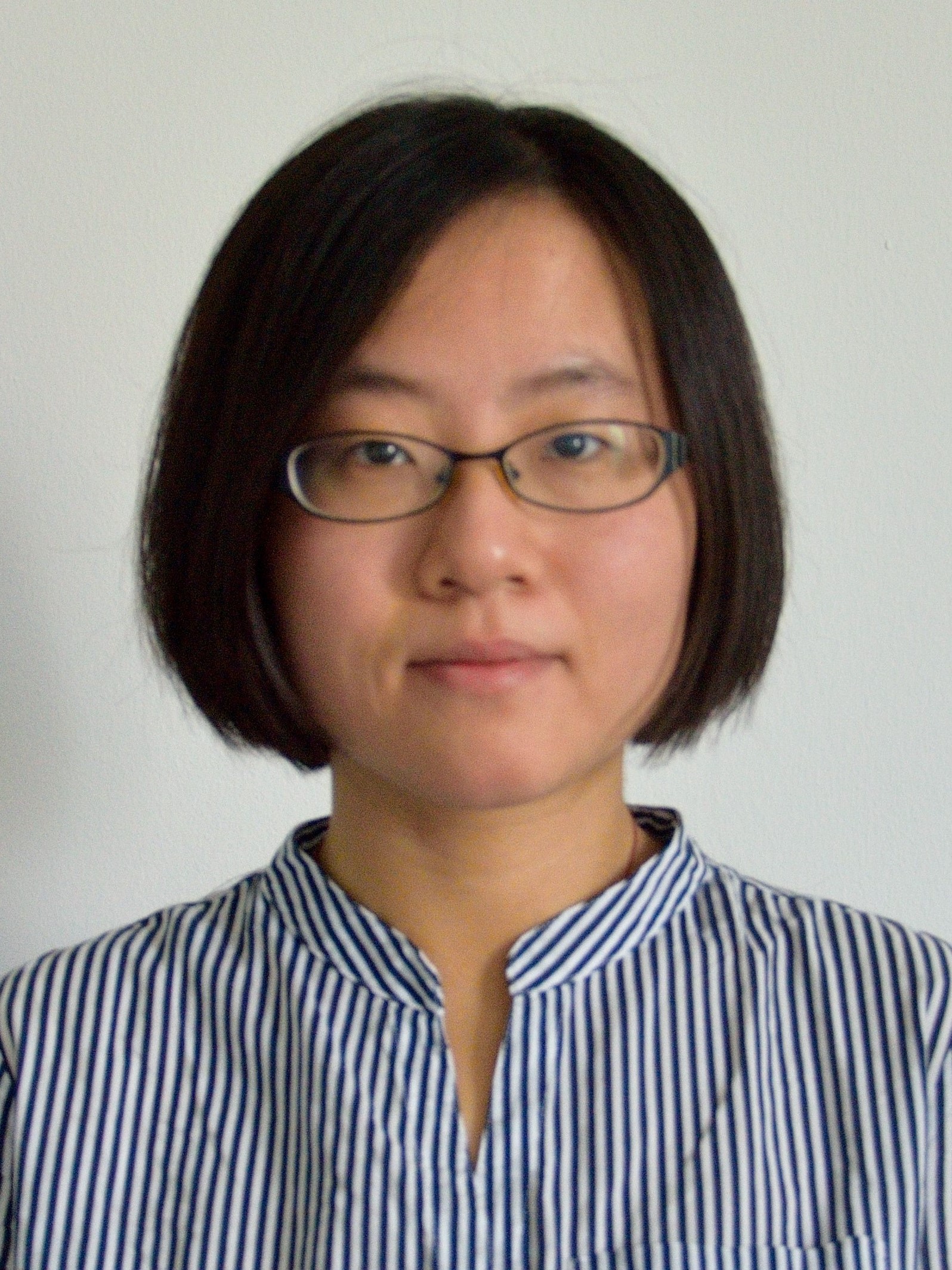}}]
 {Guichen Zhang}
 received the B.Sc. degree in remote sensing science and technology from Wuhan University, Wuhan, China, in 2014, and the M.Sc. degree in earth oriented space science and technology with the Technical University of Munich, Munich, Germany, in 2018. Since 2014, she has been a master student in photogrammetry and remote sensing with Wuhan University, Wuhan, China. Since 2018, she has been pursuing the Ph.D. degree with the German Aerospace Center, Wessling, Germany. Her research interests include image analysis and data fusion in the application of remote sensing.
 \end{IEEEbiography}

\begin{IEEEbiography}[{\includegraphics[width=1in,height=1.25in,clip,keepaspectratio]{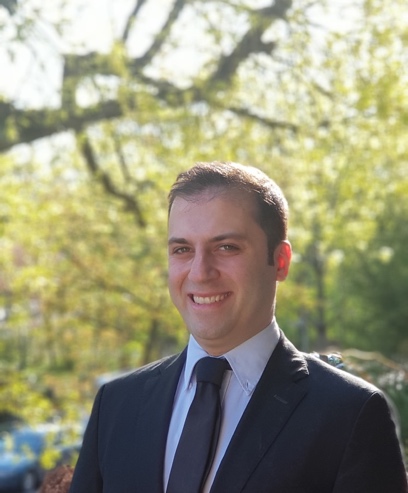}}]
{Pedram Ghamisi} (S’12--M’15--SM’12) received the B.Sc. degree in civil (survey) engineering from the Tehran South Campus of Azad University, Tehran, Iran, in 2008, the M.Sc. degree (Hons.) in remote sensing with  the K. N. Toosi University of Technology, Tehran, Iran, in 2012, and the Ph.D. degree in electrical and computer engineering with the University of Iceland, Reykjavik, Iceland, in 2015.

In 2013 and 2014, he was with the School of Geography, Planning and Environmental Management, University of Queensland, Brisbane, QLD, Australia. In 2015, he was a Post-Doctoral Research Fellow with the Technical University of Munich (TUM), Munich, Germany, and Heidelberg University, Heidelberg, Germany. From 2015 to 2018, he was a Research Scientist with the German Aerospace Center (DLR), Oberpfaffenhofen, Germany. Since 2018, he has been working as the Head of the Machine Learning Group, Helmholtz-Zentrum Dresden-Rossendorf (HZDR). He is also the CTO and the Co-Founder of VasoGnosis Inc., Milwaukee, WI, USA, where he is involved in the development of advanced diagnostic and analysis tools for brain diseases using cloud computing and deep learning algorithms. His research interests include interdisciplinary research on remote sensing and machine (deep) learning, image and signal processing, and multisensor data fusion.

Dr. Ghamisi was a recipient of the Best Researcher Award for M.Sc. students at the K. N. Toosi University of Technology in the academic year of 2010-2011, the IEEE Mikio Takagi Prize for winning the Student Paper Competition at IEEE International Geoscience and Remote Sensing Symposium (IGARSS) in 2013, the Talented International Researcher by Iran’s National Elites Foundation in 2016, the first prize of the data fusion contest organized by the Image Analysis and Data Fusion Technical Committee (IADF) of IEEE-GRSS in 2017, the Best Reviewer Prize of IEEE Geoscience and Remote Sensing Letters (GRSL) in 2017, the Alexander von Humboldt Fellowship from the Technical University of Munich, and the High Potential Program Award from HZDR. He serves as an Associate Editor for Remote Sensing and IEEE GRSL.
\end{IEEEbiography}

 \begin{IEEEbiography}[{\includegraphics[width=1in,height=1.25in,clip,keepaspectratio]{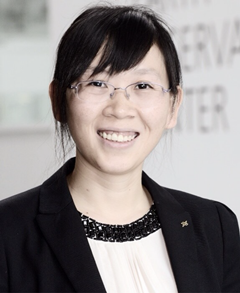}}]{Xiao Xiang Zhu}(S'10--M'12--SM'14) received the Master (M.Sc.) degree, her doctor of engineering (Dr.-Ing.) degree and her “Habilitation” in the field of signal processing from Technical University of Munich (TUM), Munich, Germany, in 2008, 2011 and 2013, respectively.
\par
She is currently the Professor for Signal Processing in Earth Observation (www.sipeo.bgu.tum.de) at Technical University of Munich (TUM) and German Aerospace Center (DLR); the head of the department ``EO Data Science'' at DLR's Earth Observation Center; and the head of the Helmholtz Young Investigator Group ``SiPEO'' at DLR and TUM. Since 2019, Zhu is co-coordinating the Munich Data Science Research School (www.mu-ds.de). She is also leading the Helmholtz Artificial Intelligence Cooperation Unit (HAICU) -- Research Field ``Aeronautics, Space and Transport". Prof. Zhu was a guest scientist or visiting professor at the Italian National Research Council (CNR-IREA), Naples, Italy, Fudan University, Shanghai, China, the University  of Tokyo, Tokyo, Japan and University of California, Los Angeles, United States in 2009, 2014, 2015 and 2016, respectively. Her main research interests are remote sensing and Earth observation, signal processing, machine learning and data science, with a special application focus on global urban mapping.

Dr. Zhu is a member of young academy (Junge Akademie/Junges Kolleg) at the Berlin-Brandenburg Academy of Sciences and Humanities and the German National  Academy of Sciences Leopoldina and the Bavarian Academy of Sciences and Humanities. She is an associate Editor of IEEE Transactions on Geoscience and Remote Sensing.

\end{IEEEbiography}
%\newpage



%\vfill


\end{document}